\PassOptionsToPackage{usenames,dvipsnames}{xcolor}
\documentclass{article}
\usepackage{arxiv}

\usepackage[utf8]{inputenc} 
\usepackage[T1]{fontenc}    
\usepackage{hyperref}       
\usepackage{url}            
\usepackage{booktabs}       
\usepackage{amsfonts}       
\usepackage{nicefrac}       
\usepackage{microtype}      
\usepackage{lipsum}		
\usepackage{graphicx}
\usepackage{natbib}
\usepackage{doi}

\usepackage{amsmath}
\usepackage{amssymb} 
\usepackage{bm}
\usepackage{bbm}
\usepackage{tikz}
\usepackage{pgfplots}
\pgfplotsset{compat=1.17}
\usepackage{fontawesome}
\usepackage{subcaption}
\usepackage{xargs} 
\usepackage{booktabs}
\usepackage{multirow}
\usepackage{array} 
\usepackage[vlined,ruled,linesnumbered]{algorithm2e}
\usepackage{ulem} 
\usepackage{contour} 
\usepackage{textcomp} 
\usepackage{pifont} 
\usepackage[switch]{lineno}  

\newcommand{\xmark}{\ding{55}}%

\contourlength{1pt}

\newcommand{\myuline}[1]{%
	\uline{\phantom{#1}}%
	\llap{\contour{white}{#1}}%
}

\makeatletter
\newcommand\footnoteref[1]{\protected@xdef\@thefnmark{\ref{#1}}\@footnotemark}
\makeatother

\newcommand{\todo}[1]{}
\renewcommand{\todo}[1]{{\color{red} TODO: {#1}}}

\SetKwComment{Comment}{$\triangleright$\ }{}

\newcommand{\vect}[1]{{\bm{\mathbf{#1}}}}

\DeclareMathOperator{\sgn}{sgn}

\usetikzlibrary{calc,positioning,shapes.geometric,arrows,fit,backgrounds}

\makeatletter
\tikzset{
	database/.style={
		cylinder,
		cylinder uses custom fill,
		cylinder body fill=yellow!50,
		cylinder end fill=yellow!50,
		shape border rotate=90,
		aspect=0.25,
		minimum height=1.2cm,
		minimum width=0.9cm,
		draw
	}
}

\title{Stateful Detection of Model Extraction Attacks}

\author{Soham Pal\\
	Indian Institute of Science, Bangalore\\
	\texttt{sohampal@iisc.ac.in}
	\And
	Yash Gupta\thanks{Work done while a graduate student at Indian Institute of Science, Bangalore.}\\
	nference\\
	\texttt{yash@nference.net}\\
	\hphantom{Indian Institute of Science, Bangalore}
	\AND
	Aditya Kanade\\
	Indian Institute of Science, Bangalore\\
	\texttt{kanade@iisc.ac.in}
	\And
	Shirish Shevade\\
	Indian Institute of Science, Bangalore\\
	\texttt{shirish@iisc.ac.in}
}

\date{}

\renewcommand{\headeright}{}
\renewcommand{\shorttitle}{Stateful Detection of Model Extraction Attacks}

\hypersetup{
pdftitle={Stateful Detection of Model Extraction Attacks},
pdfsubject={cs.LG, cs.CR},
pdfauthor={Soham~Pal, Yash~Gupta, Aditya~Kanade, Shrish~Shevade},
pdfkeywords={machine-learning-as-a-service, security and privacy, model extraction},
}

\begin{document}
\maketitle

\begin{abstract}
Machine-Learning-as-a-Service providers expose machine learning (ML) models through application programming interfaces (APIs) to developers. Recent work has shown that attackers can exploit these APIs to extract good approximations of such ML models, by querying them with samples of their choosing. We propose VarDetect, a stateful monitor that tracks the distribution of queries made by users of such a service, to detect model extraction attacks. Harnessing the latent distributions learned by a modified variational autoencoder, VarDetect robustly separates three types of attacker samples from benign samples, and successfully raises an alarm for each. Further, with VarDetect deployed as an automated defense mechanism, the extracted substitute models are found to exhibit poor performance and transferability, as intended. Finally, we demonstrate that even adaptive attackers with prior knowledge of the deployment of VarDetect, are detected by it.
\end{abstract}

\keywords{machine-learning-as-a-service \and security and privacy \and model extraction}

\section{Introduction}

The growing popularity of machine learning (ML) models has led to the rise of Machine-Learning-as-a-Service (MLaaS) offerings. Typically, MLaaS providers expose cloud ML models through black-box request-response APIs, allowing users to query the MLaaS model $g$ with an input $\vect{x}$ of their choosing, and obtain the predicted output label or probability vector $g(\vect{x})$. As many proprietary MLaaS models bill users on a pro rata basis, model architecture and weights are often withheld as a trade secret. The security and privacy of their black-box models is thus a key concern of MLaaS service providers.

Recent work by \cite{tramer} has shown that attackers with only black-box access to MLaaS models can perform \textit{model extraction attacks} to obtain a close approximation $\tilde g \approx g$. For this, an attacker generates a set of labeled pairs  $\{(\vect{x}, g(\vect{x}))\}$ by querying the \textit{MLaaS model} $g$ with samples $\vect{x}$ of its choosing. By training a new model on these pairs, the attacker obtains the extracted \textit{substitute model} $\tilde g$.

\begin{figure}[t]
	\centering
	\footnotesize
	\begin{tikzpicture}[
		>=stealth,
		node distance=3cm,
		block/.style={
			rectangle,
			aspect=0.25,
			draw,
			align=center
		}
		]	
		\node[block,minimum width=0.36\linewidth,minimum height=3.3cm] (encloseSecret) at (2,3) {};
		
		\node[database] (secret-dataset) at (0,3.5) {\Large $\mathcal{D}_\text{C}$};
		
		\node[block,fill=red!20,minimum width=2.5cm,minimum height=1.0cm,label={[label distance=0cm,align=center]above:MLaaS model},right=1.5cm of secret-dataset] (secret-model) {\huge $g(\cdot)$}; 
		
		\node [draw,fill=green!20,block,minimum height=0.5cm] (ood-check) at (2.4,2.0) {VarDetect};
		
		\node[] (trashcan) at (0,2.0) {\Huge \faWarning};
		
		\node[block,minimum width=1.5cm,minimum height=0.5cm,fill=blue!20] (X) at (7.5,2.0) {$\mathbf{x}$};
		
		\node[block,draw,minimum width=1.5cm,minimum height=0.5cm,fill=blue!20] (Y) at (secret-model.west -| X.north) {$g(\vect{x})$};
		
		\path[->,draw] (secret-dataset) edge node[above] {Train} (secret-model);
		\path[->,draw,very thick] (X.west) to node[above,pos=0.25] {Query} (ood-check.east);
		\path[->,draw] (ood-check) edge node[left] {Allow} (ood-check.north |- secret-model.south);
		\path[->,draw,very thick] (secret-model.east) to node[above,align=left,pos=0.6] (predallowed) {Prediction} (Y.west);
		\path[->,draw] (ood-check) edge node[above] {\;Raise} node[below] {\;alarm} (trashcan);
		
	\end{tikzpicture}
	\caption{Overview of our model extraction defense}
	\label{fig:model-extraction-defender}
\end{figure}
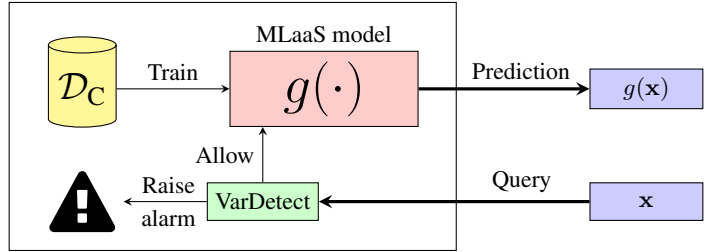

Besides the obvious threat to the pay-per-query business model of black-box MLaaS models, gradients of {$\tilde g$} can be used to generate \textit{adversarial examples} by adding human-imperceptible noise to samples so that they are misclassified by $g$, as in \cite{papernot}; or to speed up \textit{model inversion}, as in \cite{tramer}, which reveals part of the confidential dataset used to train $g$. Detecting and preventing model extraction attacks is thus key to building secure MLaaS systems.

Cloud providers of MLaaS and other services deploy various monitoring tools to detect performance and security issues. In this paper, we propose VarDetect as a stateful monitor for detecting model extraction attacks. By tracking the distribution of queries made by each user to an ML model, VarDetect raises an alarm if the distribution of user queries deviates from the expected inputs (see Figure~\ref{fig:model-extraction-defender}). The security team of the MLaaS provider can set various thresholds for detection, and map them to  alarms of increasing severity. They could then inspect the user activity to take further necessary action. To summarize, we propose VarDetect -- a monitor for the stateful detection of model extraction attacks. VarDetect has the following advantages over prior work:

\begin{itemize}
	\itemsep0em
	\item VarDetect successfully detects all three classes of attackers proposed in the literature, while allowing access to benign users.
	\item We demonstrate the effectiveness of VarDetect experimentally, across three diverse image classification tasks, wherein it reduces the accuracy and transferability of extracted models, as intended.
	\item VarDetect does not require access to attacker data.
	\item We demonstrate that VarDetect is effective against two classes of adaptive attackers, which are aware of its deployment to safeguard the MLaaS model.
\end{itemize}

\noindent We make our source code available at \myuline{\url{https://github.com/vardetect/vardetect}}.

\section{Related Work}
\label{sec:relatedwork}

\citet{deceptive-perturbation,poisoning} propose algorithms intended to reduce the performance of model extraction by perturbing the output probabilities returned by MLaaS models. While effective in reducing damage, it is not applicable when the model returns only output labels. \cite{bdpl} introduce BDPL, a Boundary Differentially Private Layer that generalizes this to binary classifiers that only return output labels, but the resulting approach causes the MLaaS provider to deliberately returning incorrect labels with a low probability. \citet{mlaaswarn} propose a model extraction monitor that is specifically applicable to decision tree classifiers, and cannot be extended to neural network classifiers. \citet{prada} propose PRADA, a defense against model extraction that is applicable only to attackers that synthesize attacker queries either through perturbation or by taking linear combinations of samples from the problem domain. Their algorithm cannot detect the most potent non-problem domain \cite{knockoff,activethief} class of attacks. While \citet{boogeyman} can defend against such attacks, they assume access to the attacker's dataset, which is unrealistic in practice. 

Our work is also closely related to the domain of anomaly detection. Prior work by \cite{magnet} and \cite{cowboy,defensegan} have leveraged autoencoders (AEs) and generative adversarial networks (GANs) respectively to protect models against adversarial examples. \citet{autoencoder-andrew} propose the use of AEs in combination with classic outlier detection methods (one-class support vector machine) for hybrid anomaly detection. Deep generative models, including variation autoencoders (VAEs) have been used in myriad other ways for outlier detection, see \cite{dlad} for further details. While anomaly detection mechanisms are typically harnessed to detect out-of-distribution samples for which the ML model may fallaciously predict a high confidence score for one of its labels, we are instead interested in safeguarding MLaaS models against model extraction.

\section{Threat Model}
\label{sec:three}
Consider a $k$-category image classification dataset $\mathcal{D}_\text{C}$. Let $g$ denote the MLaaS model trained on this \textit{confidential dataset} and $g(\vect{x})$ denote the $k$-dimensional \textit{probability vector} obtained by applying $g$ to the input $\vect{x} \in [0,1]^d$. We begin by defining the threat posed to $g$ by model extraction attackers.

\paragraph{Attack surface} Users of the MLaaS model may query the model with a sample $\vect{x}$ of their choosing. The MLaaS API, in turn, responds by returning the prediction $g(\vect{x})$ to the user. Users have no access to the weights or architecture of $g$. For the purposes of this paper, each user corresponds to a single user account of an MLaaS service, used by an individual.

\begin{figure}[t]
	\centering
	\LARGE
	\resizebox{0.55\linewidth}{!}{
		\begin{tabular}{@{}l@{}c@{}}
			\begin{tabular}{l}\bf $\mathbf{\mathcal{D}_\text{C}}$\hphantom{spacee}\\\vphantom{r}\end{tabular} & \includegraphics[width=.75\linewidth]{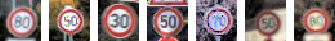}\\
			\begin{tabular}{l}\bf Syn\\\vphantom{r}\end{tabular} & \includegraphics[width=.75\linewidth]{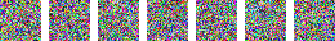}\\
			\begin{tabular}{l}\bf AdvPD\\\vphantom{r}\end{tabular} & \includegraphics[width=.75\linewidth]{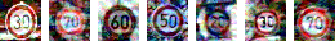}\\
			\begin{tabular}{l}\bf NPD\\\vphantom{r}\end{tabular} & \includegraphics[width=.75\linewidth]{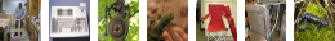}\\
			\begin{tabular}{l}\bf PD\\\vphantom{r}\end{tabular} & \includegraphics[width=.75\linewidth]{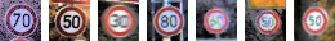}\\
			\begin{tabular}{l}\bf AltPD\\\vphantom{r}\end{tabular} & \includegraphics[width=.75\linewidth]{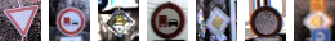}
		\end{tabular}
	}
	\caption{Representative comparison of attacker and benign samples}
	\label{fig:datasets}
\end{figure}

\paragraph{Benign user capabilities}
As we expect benign users to have access to problem domain data, we model benign users as users that query $g$ with samples from either a:
\begin{enumerate}
	\item \myuline{Problem-domain (PD) test set}: This set of samples mimics the distribution of training data used to train $g$.
	\item \myuline{Alternative problem-domain (AltPD) test set}: This contains samples belonging to classes that are not part of the training set, but are of similar nature (e.g., traffic signs that are not part of the training set, as in Figure~\ref{fig:datasets}). This is constructed using a set of held-out classes.
\end{enumerate}

\paragraph{Attacker capabilities} Attackers, much like benign users, may also query the model with inputs $\vect{x}$ of their choosing. We assume, as in prior work such as \cite{prada}, that the attacker has no or limited access to $\mathcal{D}_\text{C}$, but instead draw samples from an attacker dataset $\mathcal{D}_\text{A}$ composed of one or more of:
\begin{enumerate}
	\item \myuline{Synthetically generated samples (Syn)}: sampled from a multivariate uniform distribution, as in \cite{tramer}, 
	\item \myuline{Adversarially perturbed Problem Domain (AdvPD)}: by adding noise to a limited number of PD samples, as in \cite{papernot,prada}
	\item \myuline{Non-Problem Domain data (NPD)}: obtained by, e.g., crawling the public web for images, as in \cite{copycat,knockoff,activethief}.
\end{enumerate}
Figure~\ref{fig:datasets} visualizes samples for each of the 3 attacker and 2 benign datasets for an MLaaS model trained on the German Traffic Sign Recognition Benchmark of \cite{gtsr}.

\section{Background}
\subsection{Variational Autoencoders}
\label{sec:vae}
Variational autoencoders (VAEs), proposed originally by \cite{vae} are a class of encoder-decoder generative models. Much like autoencoders (AEs), VAEs \textit{reconstruct} their input $\vect{x}$ at the output $\vect{\tilde x}$ in a three step process:
\begin{enumerate}
	\item VAEs take as input a sample $\vect{x}$, map it through an encoder to obtain parameters of a distribution, say: $$\vect{\mu}(\vect{x}) = f_\mu(\vect{x}),\quad \vect{\sigma}(\vect{x}) = f_\sigma(\vect{x})$$ where $f_\mu$, $f_\sigma$ are neural networks.
	\item Using $\vect{\mu}, \vect{\sigma}$, a latent variable $\vect{z} \sim \mathcal{N}(\vect{\mu}, \vect{\sigma}^2)$ is drawn.
	
	When $\vect{x}$ is drawn from the same distribution as the training dataset for the VAE, $\vect{z}$ shall be incentivized to follow $\mathcal{N}(\vect{\mu}_0, \vect{\sigma}_0^2)$, where $\vect{\mu}_0$, $\vect{\sigma}_0$ are user defined parameters.
	\item Finally, a decoder neural network is used to obtain the reconstruction, $\vect{\hat x} = f_\text{dec}(\vect{z})$, desiring that $\vect{\hat x} \approx \vect{x}$.
\end{enumerate}
The training loss of the VAE is $\mathcal{L} = \mathcal{L}_\text{latent} + \rho \cdot \mathcal{L}_\text{recon}$.
\begin{equation*}
	\begin{split}
		\mathcal{L}_\text{latent} &= \mathbb{E}_{\vect{x} \sim \mathcal{D}^\text{train}} [\mathbb{KL}( \mathcal{N}(\vect{\mu}(\vect{x}), \vect{\sigma}(\vect{x})^2) || \mathcal{N}(\vect{\mu}_0, \vect{\sigma}_0^2))]\\
		\mathcal{L}_\text{recon} &= \mathbb{E}_{\vect{x} \sim \mathcal{D}^\text{train}} [\lVert \vect{x} - \vect{\hat{x}} \rVert^2]
	\end{split}
\end{equation*}
where $\mathcal{D}^\text{train}$ is the dataset on which the VAE is trained. $\mathcal{L}_\text{recon}$ is the standard autoencoder reconstruction loss. As described in step 2 above, the other loss term $\mathcal{L}_\text{latent}$ constrains $\vect{z}$ to follow the required distribution $\mathcal{N}(\vect{\mu}_0, \vect{\sigma}_0^2)$.

\subsection{Maximum Mean Discrepancy}
\noindent The maximum mean discrepancy between generating distributions of datasets $A$ and $B$ is computed as:\\
\begin{equation*}
	\Big\lVert \frac{1}{\left|A\right|} \textstyle\sum\limits_{\vect{a} \in A} \phi(\vect{a}) - \frac{1}{\left|B\right|} \textstyle\sum\limits_{\vect{b} \in B} \phi(\vect{b}) \Big\rVert_2
\end{equation*}
We use the kernel trick (Gaussian kernel, with $\sigma_g \in \{1, 5, 10$, $15, 20\}$) to replace the explicit dot product:\\
\begin{equation*}
	\phi(\vect{z})^T\phi(\vect{z'}) = K(\vect{z},\vect{z'}) = \textstyle\sum\limits_{g} \exp \Big( -\frac{\lVert \vect{z} - \vect{z}' \rVert^2}{2\sigma_g^2} \Big)
\end{equation*}
\noindent \cite{mmd} showed that, under certain conditions, the MMD tends to zero asymptotically if the generating distributions of $A$ and $B$ are the same.
As shown in Algorithm~\ref{alg:vardetect}, subsampling may be used to reduce the computation cost, drawing $M$ samples each of size $N$ from $A$ and $B$. In our experiments, we set $M = 100$ and subsample size $N = 20$.

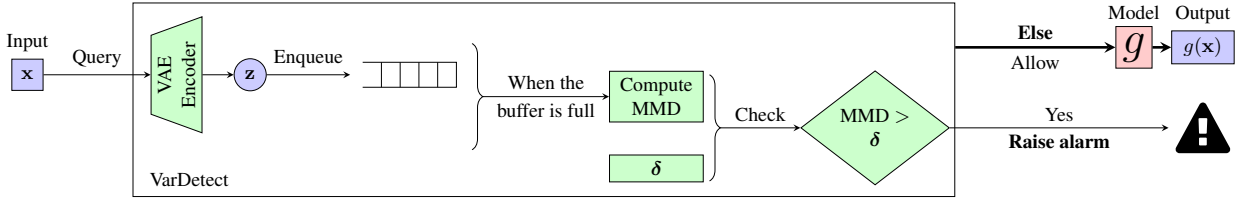
\begin{figure*}[t]
	\centering
	\footnotesize
	\resizebox{\linewidth}{!}{
		\begin{tikzpicture}[
			>=stealth,
			node distance=3cm,
			block/.style={
				rectangle,
				aspect=0.25,
				draw,
				align=center
			}
			]
			\linespread{0.75}
			
			\node[block,minimum width=0.5cm,minimum height=0.5cm,fill=blue!20,opacity=0] (X) at (2.8,-3.6) {$\mathbf{x}$};
			
			\node[block,left=0.7cm of X,minimum width=0.5cm,minimum height=0.5cm,fill=blue!20,label={[label distance=0cm,align=center]above:Input}] (actualX) {$\mathbf{x}$};
			
			\node[trapezium,fill=green!20,trapezium left angle=70, trapezium right angle=70,align=center,draw,rotate=90,right=0.5cm of X,anchor=north,font=\linespread{1.1}\selectfont] (vae) {VAE\\Encoder};
			
			\path[->,draw] (actualX) to node[above] {Query} (vae);
			
			\node[circle,draw,minimum width=0.5cm,minimum height=0.5cm,fill=blue!20,right=0.5cm of vae.south] (Z) {$\mathbf{z}$};
			
			\path[->,draw] (vae) -- (Z);
			
			\draw ($(Z) + (1.8cm,0.2cm)$) -- ++(1.5cm,0) -- ++(0,-0.4cm) -- ++(-1.5cm,0);
			\foreach \i in {1,...,4}
			\draw ($(Z) + (1.8cm+\i*0.3cm,0.2cm)$) -- +(0,-0.4cm);
			
			\path[->,draw] (Z) edge node[above] {Enqueue} ($(Z) + (1.6cm,0)$);
			
			\node[database,opacity=0] at ($(Z) + (2.55cm,-1.5cm)$) (secret-dataset) {\Large $\mathcal{D}_\text{C}$};
			
			\draw [decorate,decoration={brace,amplitude=5pt},xshift=4pt,yshift=0pt]
			($(Z) + (3.55cm,0.5cm)$) -- ($(Z) + (3.55cm,-1.2cm)$);
			
			\node[rectangle,fill=green!20,align=center,draw,minimum width=1.5cm,font=\linespread{1.1}\selectfont] at ($(Z) + (6.5cm,-0.35cm)$) (calc-mmd) {Compute\\MMD};
			
			\path ($(Z) + (3.55cm,-0.35cm) + (5pt,0)$) -- node[above]{When the} (calc-mmd);
			\path[->,draw] ($(Z) + (3.55cm,-0.35cm) + (5pt,0)$) -- node[below]{buffer is full} (calc-mmd);
			
			\node[rectangle,fill=green!20,align=center,draw,minimum width=1.5cm,opacity=1] at (calc-mmd |- secret-dataset) (musigma) {$\bm{\delta}$};
			
			
			\draw [decorate,decoration={brace,amplitude=5pt},xshift=4pt,yshift=0pt]
			($(Z) + (7.35cm,0cm)$) -- ($(Z) + (7.35cm,-1.7cm)$);
			
			\node[diamond,fill=green!20,draw,aspect=1.3,align=center,font=\linespread{1.1}\selectfont] at ($(Z) + (10cm,-0.85cm)$) (mmd-check) {MMD $>$ \\ $\bm{\delta}$};
			
			\path[->,draw] ($(Z) + (7.35cm,-0.85cm) + (5pt,0)$) -- node[above] {Check} (mmd-check);
			
			\node[right=3.5cm of mmd-check] (block) {\Huge \faWarning};
			
			\node[block,above=0.5cm of block,minimum width=1cm,minimum height=0.5cm,fill=blue!20,label={[label distance=0cm,align=center]above:Output}] (allow) {$g(\mathbf{x})$};
			
			\node[block,left=0.3cm of allow,minimum width=0.5cm,minimum height=0.5cm,fill=red!20,label={[label distance=0cm,align=center]above:Model}] (M) {\huge $g$};
			
			\path[->,draw] (mmd-check) to node[above] {Yes} node[below] {\bf Raise alarm} (block);
			
			\node[block,minimum width=13.15cm,minimum height=3.1cm,label={[xshift=47pt,yshift=14pt]south west:VarDetect}] (encloseSecret) at (9.85,-4.0) {};
			
			\path[->,draw,very thick] (encloseSecret.east |- M) to node[above] {\bf Else} node[below] {Allow} (M.west);
			\path[->,draw,very thick] (M.east) -- (allow.west);
			
	\end{tikzpicture}}
	\caption{The VarDetect framework: obtaining embeddings, MMD computation and thresholding. Conditional paths not drawn lead to \textbf{Else}.}
	\label{fig:inner-workings}
\end{figure*}

\begin{algorithm}[t]
	\newcommand\mycommfont[1]{\footnotesize\textcolor{blue}{#1}}
	\SetCommentSty{mycommfont}
	\SetAlgoLined
	\SetKwFunction{mmd}{mmd}
	
	\SetKwFunction{algo}{algo}\SetKwFunction{mmd}{mmd}
	\SetKwProg{mymmd}{Procedure}{}{}
		\mymmd{\mmd{A, B}}{
		\nl \For{$i = 1, 2, \dots M$}{
			$A_i \gets \{a \sim A: j = 1, 2, \dots N\}$  \Comment*[r]{Subsample from both sets uniformly at random}
			$B_i \gets \{b \sim B: j = 1, 2, \dots N\}$\;	
		}
		\nl $J \gets \bigcup\limits_{i=1}^M \Big\{\Big\lVert \frac{1}{\left|A_i\right|} \textstyle\sum\limits_{\vect{a} \in A_i} \phi(\vect{a}) - \frac{1}{\left|B_i\right|} \textstyle\sum\limits_{\vect{b} \in B_i} \phi(\vect{b}) \Big\rVert_2$\Big\}\;
		\nl $\vect{\mu} \gets \frac{1}{\left| J_i \right|} \sum_i J_i$\;
		\nl \KwRet $\vect{\mu}$\;}
	
	\setcounter{AlgoLine}{0}
	\SetKwProg{myalg}{Algorithm}{}{}
	\myalg{VarDetect-Precompute}{
		\For{$\vect{x}^i \in \mathcal{D}_\text{C}$}{
			$\nu^i \sim U(0,1)$ \Comment*[r]{Create outlier dataset}
			$\vect{n}^i \sim \mathcal{N}(\nu^i, 1)^d$\;
		}
		\nl $\mathcal{D}_\text{O} \gets \cup_i \{\min(\max(\tau^i \cdot \vect{x}^i + (1 - \tau^i) \cdot \vect{n}^i, \vect{0}), \vect{1})\}$\;
		\nl $f_\mu, f_\sigma \gets$ Train modified VAE on $\mathcal{D}_\text{C}$ and $\mathcal{D}_\text{O}$\;
		\nl $u = \{f_\mu(\vect{x}) : \vect{x} \in \mathcal{D^\text{train}_\text{C}}\}$\;}{}
	
	\setcounter{AlgoLine}{0}
	\myalg{VarDetect-Process-User}{
		$\mathcal{H} \gets \textsc{Queue}()$\;
		\For{input $\vect{x}$ from user}{
			$\mathcal{H}.\textsc{enqueue}(\{\vect{x}\})$\;
			\If{$\mathcal{H}.\textsc{length} > m$}{
				$\mathcal{H}.\textsc{dequeue}()$\;
				\lIf{\mmd{u, $\mathcal{H}$} $ > \bm{\delta}$}{
					raise alarm
				}
			}
		}}{}

	\caption{VarDetect Framework}
	\label{alg:vardetect}
\end{algorithm}

\section{The Proposed VarDetect Monitor}
We design VarDetect to continuously monitor the distribution of queries to $g$ from each user. As shown in Figure~\ref{fig:inner-workings}, VarDetect buffers incoming queries, and stores them in a queue of some fixed size, $m$. Each query is added to this queue. If a user's queue is full, the oldest query is removed and the buffered queries are checked against the distribution of training queries by computing the MMD between their latent distributions. Whenever the observed MMD exceeds a specified threshold $\bm{\delta}$, an alarm is raised, pointing the security team to the suspicious user and their activity. Algorithm~\ref{alg:vardetect} details the procedure for outlier dataset generation and detection.

\begin{enumerate}
	\item In lieu of requiring access to attacker samples as in \cite{boogeyman}, we construct an outlier dataset $\mathcal{D}_\text{O}$ to contrast $\mathcal{D}_\text{C}$ against. An immediate possibility is to use adversarial perturbations on the confidential dataset to construct the outlier dataset; however, first, it has been shown by \cite{not_easily_detected} that it is easy to fool a network trained to detect adversarial examples; and second, we wish to avoid mimicking the AdvPD attacker specifically. We demonstrate experimentally that by building $\mathcal{D}_\text{O}$ by simply adding noise to PD samples, not only does the resulting VAE learn to separate out all (Syn, AdvPD and NPD) attackers, but that it is also more resilient to evasion attacks (as we shall show in Section~\ref{sec:evasion}). Future work may explore other constructions of $\mathcal{D}_\text{O}$.
	
	\item VarDetect trains a class-conditional variational autoencoder to map $\mathcal{D}_\text{C}$ and $\mathcal{D}_\text{O}$ to distinct regions in latent space, by modifying the objective function:
	\begin{equation*}
		\begin{split}
			\mathcal{L}_\text{latent} = \mathbb{E}_{\vect{x} \sim \mathcal{D}_\text{C}} &\big[ \mathbb{KL}\big( \mathcal{N}(\vect{\mu}(\vect{x}), \vect{\sigma}(\vect{x})^2) ||  \mathcal{N}(\vect{\vect{\mu}_\text{C}}, \vect{\sigma_\text{C}^2})\big) \big]\\
			+\; \mathbb{E}_{\vect{x} \sim \mathcal{D}_\text{O}} &\big[ \mathbb{KL}\big( \mathcal{N}(\vect{\mu}(\vect{x}), \vect{\sigma}(\vect{x})^2) ||  \mathcal{N}(\vect{\vect{\mu}_\text{O}}, \vect{\sigma_\text{O}^2})\big) \big]
		\end{split}
	\end{equation*}
	where $\vect{\mu}_\text{C}, \vect{\sigma}_\text{C}$ and $\vect{\mu}_\text{O}, \vect{\sigma}_\text{O}$ are chosen appropriately to separate the mappings of confidential and outlier samples in latent space. We note that the 2 classes modeled by this VAE are benign and outlier (at test time, attacker) samples, and do not correspond to the classes of the original dataset used for training the model $g$.
	
	\item VarDetect is stateful by design: at test time, VarDetect matches distribution by computing the MMD between the latent mapping of a user's query history and those of $\mathcal{D}_\text{C}$ training samples. We expect benign users to generate a lower test-time MMD than attackers, as their samples should more closely resemble $\mathcal{D}_\text{C}$. An alarm is raised when the provider-specified threshold $\bm{\delta}$ is crossed.
\end{enumerate} 

Note that unlike in the work of \citet{boogeyman} or the anomaly detection works discussed in Section~\ref{sec:relatedwork}, we are not interested in detecting individual suspicious queries for two reasons: First, raising alarms for each such query can overwhelm the security team of the MLaaS provider. Second, even benign users may occassionally make queries that can be judged to be outliers. Such single outliers are not a security threat from the perspective of model extraction -- we therefore look for sustained malicious behaviors instead.

If and when an alarm is determined to be false by the security team (e.g., due to data drift), the provider may consider adding the user's samples to the VAE training dataset and retraining it, to further reduce the incidence of false alarms.

\section{Experimental Setup}

A brief summary of our experimental setup follows, with further details made available in our public repository.

\subsection{Network Architectures}
\paragraph{Image Classifiers}
We use a convolutional neural network for the MLaaS and substitute models, having 3 blocks of conv\textrightarrow batch\_norm\textrightarrow conv\textrightarrow batch\_norm\textrightarrow pool layers. The conv and pool kernels are $3\times3$ and $2\times2$. The number of filters in the 3 blocks are 32, 64 and 128, and all activations are ReLU. The final volume is flattened and passed through an output projection layer with a softmax activation.

\paragraph{VAE Encoder}
The VAE encoder passes the input $\vect{x}$ through 4 conv layers with 32, 64, 128 and 256 filters of size $4 \times 4$. The resulting volume is flattened, and projected through a dense layer to a 512-dimensional vector. This vector is passed through 2 separate feedforward networks to produce the mean $\vect{\mu}$ and standard deviation $\vect{\sigma}$ vectors. All activations are ReLU. Finally, $\vect{z}$ is sampled from $\mathcal{N}(\vect{\mu}, \vect{\sigma}^2)$.

\paragraph{VAE Decoder}
$\vect{z}$ is passed through a dense layer of size $512$. The resulting vector is reshaped into a one-dimensional volume, and passed through 4 deconv layers, with 256, 128, 64 and 32 filters of size $4 \times 4$. All activations are ReLU. The final volume is passed through a similar deconvolution layer (producing the required number of channels) with a sigmoid activation to obtain the final reconstructed image.

\subsection{Datasets}
We use a wide range of image classification datasets as confidential datasets, namely: the simple grayscale 10-class Fashion-MNIST (F-MNIST) of \cite{fmnist}, the color 10-class Street View House Numbers (SVHN) of \cite{svhn} and the color 43-class German Traffic Sign Recognition (GTSR) benchmark of \cite{gtsr}.
Our NPD attacker uses ImageNet samples of \cite{imagenet} as a proxy for non-problem domain data, as in \cite{activethief}.

\subsection{Hyperparameters}
\paragraph{Network and Loss}
Dropout is applied at a rate of 0.2 on all layers with ReLU activations. The VAEs are trained with a 32-dimensional latent variable. The loss is configured with a reconstruction loss multiplier $\rho = 0.5$. We use the means $\vect{\vect{\mu}_\text{C}} = \vect{0}$, $\vect{\vect{\mu}_\text{O}} = 5 \cdot \vect{1}$ and uncorrelated unit-variance $\vect{\sigma_\text{C}} = \vect{\sigma_\text{O}} = \vect{1}$, where $\vect{0}$ and $\vect{1}$ are zero and all-ones vectors, as before. The Adam optimizer of \cite{adam} is used.

\paragraph{Training Hyperparameters}
Convolutional layers in the classifiers use the He initializer, and all other layers use a Glorot initializer. The VAE is trained for up to 500 epochs, until convergence. For the Syn and AdvPD attacks, the substitute model is trained for 50 and 100 epochs respectively. For NPD attacks, training is performed up to 1000 epochs, employing early stopping with a patience of 10 epochs, validating on the $F_1$ measure on the validation set.

\section{Experimental Results}

We extensively evaluate VarDetect against a suite of 12 attackers. In the main paper, we select 3 representative attacks: uniform retraining of \citet{tramer} (Syn), the JSMA attack of \citet{papernot} (AdvPD) and the ensemble strategy of \citet{activethief} (NPD). Extended results for the remaining attacks are presented in Appendix~\ref{apd:first}.

\subsection{Comparison of Attackers and Benign Users}

We first study how the MMD values evolve for attackers and benign users. The AltPD case (as defined in Section~\ref{sec:three}) requires that the attacker has access to a set of classes that are not used during training. To ensure uniformity, we hold out half of the classes, and train the classifiers and VAEs on the remaining half. All the attackers and benign users are then evaluated on the same models. The same setting is used for plotting learned latent representations below. Other than these experiments, we use the original splits of the datasets.


\begin{figure*}[t]
	\centering
	\begin{subfigure}{0.32\textwidth}
		\begin{tikzpicture}
			\begin{axis}[
				legend style={at={(0.5,-0.55)},
					anchor=north,legend columns=-1,font=\tiny},
				axis x line=left,
				axis y line=left,
				bar width=0.2cm,
				yticklabels={},
				enlarge y limits=0.10,
				extra y ticks={0.25},
				extra y tick style={
					grid style={
						black,
						dashed,
						/pgfplots/on layer=axis foreground,
					},
				},
				width=\linewidth,
				height=4cm
				]
				\addplot[very thick,color=Maroon,dotted] table[x=x,y=y,col sep=space] {latex_data/vae/mmd/fashion-img-1337.txt};
				\addplot[very thick,color=Purple,loosely dotted] table[x=x,y=y,col sep=space] {latex_data/vae/mmd/fashion-unif-1337.txt};
				\addplot[thick,color=orange] table[x=x,y=y,col sep=space] {latex_data/vae/mmd/fashion-test-1337.txt};
				\addplot[thick,color=ForestGreen] table[x=x,y=y,col sep=space] {latex_data/vae/mmd/fashion-altpd-1337.txt};
				\addplot[thick,color=Cerulean,dashed] table[x=x,y=y,col sep=space] {latex_data/vae/mmd/fashion-pap-1337.txt};
			\end{axis}
		\end{tikzpicture}
		\caption{F-MNIST dataset}
	\end{subfigure}
	\begin{subfigure}{0.32\textwidth}
		\begin{tikzpicture}
			\begin{axis}[
				legend style={at={(0.5,-0.55)},
					anchor=north,legend columns=-1,font=\tiny},
				axis x line=left,
				axis y line=left,
				bar width=0.2cm,
				enlarge y limits=0.10,
				yticklabels={},
				extra y ticks={0.25},
				extra y tick style={
					grid style={
						black,
						dashed,
						/pgfplots/on layer=axis foreground,
					},
				},
				width=\linewidth,
				height=4cm
				]
				\addplot[very thick,color=Maroon,dotted] table[x=x,y=y,col sep=space] {latex_data/vae/mmd/gtsr-img-1337.txt};
				\addplot[very thick,color=Purple,loosely dotted] table[x=x,y=y,col sep=space] {latex_data/vae/mmd/gtsr-unif-1337.txt};
				\addplot[thick,color=orange] table[x=x,y=y,col sep=space] {latex_data/vae/mmd/gtsr-test-1337.txt};
				\addplot[thick,color=ForestGreen] table[x=x,y=y,col sep=space] {latex_data/vae/mmd/gtsr-altpd-1337.txt};
				\addplot[thick,color=Cerulean,dashed] table[x=x,y=y,col sep=space] {latex_data/vae/mmd/gtsr-pap-1337.txt};
			\end{axis}
		\end{tikzpicture}
		\caption{GTSR dataset}
	\end{subfigure}
	\begin{subfigure}{0.32\textwidth}
		\begin{tikzpicture}
			\begin{axis}[
				legend style={at={(0.5,-0.55)},
					anchor=north,legend columns=-1,font=\tiny},
				axis x line=left,
				axis y line=left,
				bar width=0.2cm,
				enlarge y limits=0.10,
				yticklabels={},
				ytick={},
				extra y ticks={0.25},
				extra y tick style={
					grid style={
						black,
						dashed,
						/pgfplots/on layer=axis foreground,
					},
				},
				width=\linewidth,
				height=4cm
				]
				\addplot[very thick,color=Maroon,dotted] table[x=x,y=y,col sep=space] {latex_data/vae/mmd/svhn-img-1337.txt};\label{img_mmd}
				\addplot[very thick,color=Purple,loosely dotted] table[x=x,y=y,col sep=space] {latex_data/vae/mmd/svhn-unif-1337.txt};\label{unif_mmd}
				\addplot[thick,color=orange] table[x=x,y=y,col sep=space] {latex_data/vae/mmd/svhn-test-1337.txt};\label{benign_mmd}
				\addplot[thick,color=ForestGreen] table[x=x,y=y,col sep=space] {latex_data/vae/mmd/svhn-altpd-1337.txt};\label{altpd_mmd}
				\addplot[thick,color=Cerulean,dashed] table[x=x,y=y,col sep=space] {latex_data/vae/mmd/svhn-pap-1337.txt};\label{pap_mmd}
			\end{axis}
		\end{tikzpicture}
		\caption{SVHN dataset}
	\end{subfigure}
	\caption{The change in MMD (Y axis), as computed by VarDetect over time: for the PD test set \ref{benign_mmd}, AltPD test set \ref{altpd_mmd}, Syn \ref{unif_mmd}, AdvPD \ref{pap_mmd} and NPD \ref{img_mmd} attackers. The X-axis indicates the cumulative number of queries fired by each user at the time of MMD computation. The process begins only after the query buffer for each user is full, i.e., $\mathcal{H}.\text{\textsc{length}} > m$ (we use $m = 100$ in our experiments).}
	\label{fig:mmdovertime}
\end{figure*}
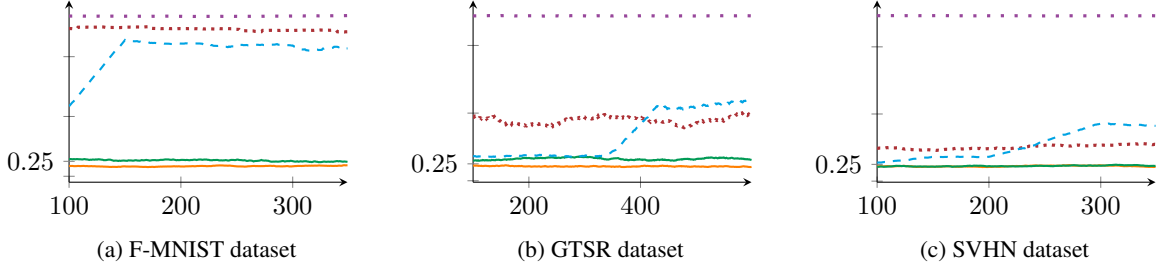

\paragraph{MMD over Time}
As shown in Figure~\ref{fig:mmdovertime}, VarDetect maps both PD and AltPD benign user datasets to low MMD values. Thus, VarDetect admits both PD and AltPD benign users; the latter contains classes not seen during training.

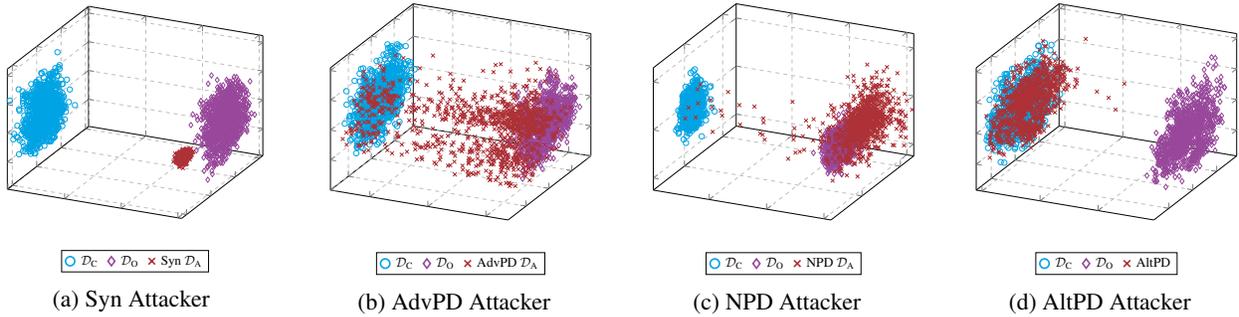
\begin{figure*}[t]
	\centering
	\begin{subfigure}{0.22\textwidth}
		\centering
		\resizebox{\linewidth}{!}{
			\begin{tikzpicture}
				\begin{axis}[x dir=reverse,grid=major,grid style={dashed, gray!50},xticklabels={},yticklabels={},zticklabels={},legend image post style={scale=1.75,very thick},
					legend style={/tikz/every even column/.append style={column sep=0.2cm},at={(0.5,-0.15)},
						anchor=north,legend columns=-1}]
					\addplot3+[only marks,mark=o,color=Cerulean] table {latex_data/vae/fashionmnist/pca/new/pos_uniform_neg-pos.txt};
					\addlegendentry{$\phantom{.}\mathcal{D}_\text{C}$}
					\addplot3+[only marks,mark=diamond,color=Purple] table {latex_data/vae/fashionmnist/pca/new/pos_uniform_neg-neg.txt};
					\addlegendentry{$\phantom{.}\mathcal{D}_\text{O}$}
					\addplot3+[only marks,mark=x,color=Maroon] table {latex_data/vae/fashionmnist/pca/new/pos_uniform_neg-uniform.txt};
					\addlegendentry{\phantom{.}Syn $\mathcal{D}_\text{A}$}
				\end{axis}
		\end{tikzpicture}}
		\caption{Syn Attacker}
	\end{subfigure}\hfill
	\begin{subfigure}{0.22\textwidth}
		\centering
		\resizebox{\linewidth}{!}{
			\begin{tikzpicture}
				\begin{axis}[x dir=reverse,grid=major,grid style={dashed, gray!50},xticklabels={},yticklabels={},zticklabels={},legend image post style={scale=1.75,very thick},
					legend style={/tikz/every even column/.append style={column sep=0.2cm},at={(0.5,-0.15)},
						anchor=north,legend columns=-1}]
					\addplot3+[only marks,mark=o,color=Cerulean] table {latex_data/vae/fashionmnist/pca/new/pos_jbda_neg-pos.txt};
					\addlegendentry{\phantom{.}$\mathcal{D}_\text{C}$}
					\addplot3+[only marks,mark=diamond,color=Purple] table {latex_data/vae/fashionmnist/pca/new/pos_jbda_neg-neg.txt};
					\addlegendentry{\phantom{.}$\mathcal{D}_\text{O}$}
					\addplot3+[only marks,mark=x,color=Maroon] table {latex_data/vae/fashionmnist/pca/new/pos_jbda_neg-seed.txt};
					\addplot3+[only marks,mark=x,color=Maroon] table {latex_data/vae/fashionmnist/pca/new/pos_jbda_neg-aug.txt};
					\addlegendentry{\phantom{.}AdvPD $\mathcal{D}_\text{A}$}
				\end{axis}
		\end{tikzpicture}}
		\caption{AdvPD Attacker}
	\end{subfigure}\hfill
	\begin{subfigure}{0.22\textwidth}
		\centering
		\resizebox{\linewidth}{!}{
			\begin{tikzpicture}
				\begin{axis}[x dir=reverse,grid=major,grid style={dashed, gray!50},xticklabels={},yticklabels={},zticklabels={},legend image post style={scale=1.75,very thick},
					legend style={/tikz/every even column/.append style={column sep=0.2cm},at={(0.5,-0.15)},
						anchor=north,legend columns=-1}]
					\addplot3+[only marks,mark=o,color=Cerulean] table {latex_data/vae/fashionmnist/pca/new/pos_imagenet_neg-pos.txt};
					\addlegendentry{\phantom{.}$\mathcal{D}_\text{C}$}
					\addplot3+[only marks,mark=diamond,color=Purple] table {latex_data/vae/fashionmnist/pca/new/pos_imagenet_neg-neg.txt};
					\addlegendentry{\phantom{.}$\mathcal{D}_\text{O}$}
					\addplot3+[only marks,mark=x,color=Maroon] table {latex_data/vae/fashionmnist/pca/new/pos_imagenet_neg-imagenet.txt};
					\addlegendentry{\phantom{.}NPD $\mathcal{D}_\text{A}$}
				\end{axis}
		\end{tikzpicture}}
		\caption{NPD Attacker}
	\end{subfigure}\hfill
	\begin{subfigure}{0.22\textwidth}
		\centering
		\resizebox{\linewidth}{!}{
			\begin{tikzpicture}
				\begin{axis}[grid=major,grid style={dashed, gray!50},xticklabels={},yticklabels={},zticklabels={},legend image post style={scale=1.75,very thick},
					legend style={/tikz/every even column/.append style={column sep=0.2cm},at={(0.5,-0.15)},
						anchor=north,legend columns=-1}]
					\addplot3+[only marks,mark=o,color=Cerulean] table {latex_data/vae/fashionmnist/pca/new/pos_altpd_neg-pos.txt};
					\addlegendentry{\phantom{.}$\mathcal{D}_\text{C}$}
					\addplot3+[only marks,mark=diamond,color=Purple] table {latex_data/vae/fashionmnist/pca/new/pos_altpd_neg-neg.txt};
					\addlegendentry{\phantom{.}$\mathcal{D}_\text{O}$}
					\addplot3+[only marks,mark=x,color=Maroon] table {latex_data/vae/fashionmnist/pca/new/pos_altpd_neg-altpd.txt};
					\addlegendentry{\phantom{.}AltPD}
				\end{axis}
		\end{tikzpicture}}
		\caption{AltPD Attacker}
	\end{subfigure}

	\caption{Projection of the latent vectors $\vect{z}$ into 3D (via a PCA projection) for F-MNIST confidential dataset $\mathcal{D}_\text{C}$, outlier dataset $\mathcal{D}_\text{O}$ and various attacker dataset $\mathcal{D}_\text{A}$ samples -- revealing that encodings of attacker samples lie closer to outlier samples.}
	\label{fig:projection}
\end{figure*}

We also observe that the MMD values for Syn, AdvPD and NPD attackers are clearly well-separated, and much higher than that for benign users. Thus, using an appropriate threshold, VarDetect can be deployed to detect the attackers without unnecessarily flagging benign users to the security team.

\begin{table}[t]
	\centering
	\caption{Detection at different thresholds: \xmark\; indicates that an alarm has not been raised.}
	\label{tab:detection}
	\vspace{1em}
	
	\begin{subtable}{\linewidth}
		\centering
		\vspace{1em}
		\begin{tabular}{lccccc}
			\toprule
			Threshold & Syn & AdvPD & NPD & AltPD & PD \\
			\midrule
			0.00 & Alarm & Alarm & Alarm & Alarm & Alarm \\
			0.25 & Alarm & Alarm & Alarm & Alarm & \xmark \\
			0.50 & Alarm & Alarm & Alarm & \xmark & \xmark \\
			1.00 & Alarm & Alarm & Alarm & \xmark & \xmark \\
			1.50 & Alarm & Alarm & Alarm & \xmark & \xmark \\
			2.50 & \xmark & \xmark & \xmark & \xmark & \xmark \\
			\bottomrule
		\end{tabular}
		\caption{F-MNIST}
	\end{subtable}\\
	\begin{subtable}{\linewidth}
		\centering
		\vspace{1em}
		\begin{tabular}{lccccc}
			\toprule
			Threshold & Syn & AdvPD & NPD & AltPD & PD \\
			\midrule
			0.00 & Alarm & Alarm & Alarm & Alarm & Alarm \\
			0.25 & Alarm & Alarm & Alarm & Alarm & \xmark \\
			0.50 & Alarm & Alarm & Alarm & \xmark & \xmark \\
			1.00 & Alarm & Alarm & \xmark & \xmark & \xmark \\
			1.50 & Alarm & \xmark & \xmark & \xmark & \xmark \\
			2.50 & \xmark & \xmark & \xmark & \xmark & \xmark \\
			\bottomrule
		\end{tabular}
		\caption{GTSR}
	\end{subtable}\\
	\begin{subtable}{\linewidth}
		\centering
		\vspace{1em}
		\begin{tabular}{lccccc}
			\toprule
			Threshold & Syn & AdvPD & NPD & AltPD & PD \\
			\midrule
			0.00 & Alarm & Alarm & Alarm & Alarm & Alarm \\
			0.25 & Alarm & Alarm & Alarm & \xmark & \xmark \\
			0.50 & Alarm & Alarm & Alarm & \xmark & \xmark \\
			1.00 & Alarm & Alarm & \xmark & \xmark & \xmark \\
			1.50 & Alarm & \xmark & \xmark & \xmark & \xmark \\
			2.50 & \xmark & \xmark & \xmark & \xmark & \xmark \\
			\bottomrule
		\end{tabular}
		\caption{SVHN}
	\end{subtable}
\end{table}

In Table~\ref{tab:detection}, we summarize whether an alarm is raised or not for each of the benign and attacker datasets for different threshold values. In general, by increasing the threshold, VarDetect is made gradually more forgiving: in the order PD, then AltPD, then NPD, then AdvPD and finally Syn. Thus, Syn data is rejected with extreme ease, while NPD is the hardest to reject. Using a threshold of 0.5 or greater allows us to admit PD and AltPD, while detecting all the attacker cases. A threshold of 0.25 is sufficient to allow PD benign users.

\paragraph{Learned Latent Space Representations} To understand the success of VarDetect in distiguishing between benign users and attackers, we inspect their latent representations. In Figure~\ref{fig:projection}, we show PCA projections of VAE encodings of the confidential dataset $\mathcal{D}_\text{C}$ and outlier dataset $\mathcal{D}_\text{O}$, along with the three attackers and the AltPD benign user. These are shown for the F-MNIST dataset.

We observe that our modified VAE learns to clearly separate out $\mathcal{D}_\text{C}$ and $\mathcal{D}_\text{O}$ datasets. Further, each of the attacker datasets are mapped away from $\mathcal{D}_\text{C}$ and towards $\mathcal{D}_\text{O}$. The AltPD benign data points however get mapped closer to $\mathcal{D}_\text{C}$ than $\mathcal{D}_\text{O}$. This results in the MMD values for attackers being higher than MMD values for benign users.

\begin{table}[t]
	\centering
	
	\caption{The test set accuracy (\%) of substitute models obtained by model extraction attackers, when the MLaaS model is defended using VarDetect, compared to when it is not (lower is better), for attackers with a budget of 100,000 queries.}
	\label{tab:accuracy}
	\vspace{1em}
	
	\begin{tabular}{lrrrrrr}
		\toprule
		& \multicolumn{3}{c}{Undefended} & \multicolumn{3}{c}{Defended} \\ \cmidrule(lr){2-4} \cmidrule(lr){5-7}
		& \multicolumn{1}{c}{Syn} & \multicolumn{1}{c}{AdvPD} & \multicolumn{1}{c}{NPD} & \multicolumn{1}{c}{Syn} & \multicolumn{1}{c}{AdvPD} & \multicolumn{1}{c}{NPD}\\
		\midrule
		F-MNIST & 25.33 & 84.66 & 81.10 & \bf 7.56 & \bf 75.50 & \bf 12.24 \\
		GTSR & 69.10 & 90.44 & 93.96 & \bf 5.19 & \bf 54.07 & \bf 6.74 \\
		SVHN & 50.71 & 71.12 & 92.57 & \bf 14.57 & \bf 41.38 & \bf 15.67 \\
		\bottomrule
	\end{tabular}
\end{table}

\begin{table}[t]
	\centering
	
	\caption{The test set transferability (\%) of substitute models obtained by model extraction attackers, when the MLaaS model is defended using VarDetect, compared to when it is not (lower is better), for attackers with a budget of 100,000 queries.}
	\label{tab:transferability}
	\vspace{1em}
	
	\begin{tabular}{lrrrrrr}
		\toprule
		& \multicolumn{3}{c}{Undefended} & \multicolumn{3}{c}{Defended} \\ \cmidrule(lr){2-4} \cmidrule(lr){5-7}
		& \multicolumn{1}{c}{Syn} & \multicolumn{1}{c}{AdvPD} & \multicolumn{1}{c}{NPD} & \multicolumn{1}{c}{Syn} & \multicolumn{1}{c}{AdvPD} & \multicolumn{1}{c}{NPD}\\
		\midrule
		F-MNIST & 60.74 & 68.73 & 78.74 & \bf 42.35 & \bf 63.13 & \bf 55.79 \\
		GTSR & 84.51 & 80.81 & 94.56 & \bf 13.61 & \bf 68.53 & \bf 37.94 \\
		SVHN & 82.51 & 82.30 & 91.30 & \bf 31.54 & \bf 53.52 & \bf 26.87 \\
		\bottomrule
	\end{tabular}
	
\end{table}
\subsection{Performance of Extracted Models (Accuracy and Transferability)}
\label{sec:acctrans}

We now deploy VarDetect as an automated defense mechanism for the detection of all three attacks, using the original splits of the datasets (PD benign users only). We set the threshold value to $\delta = 0.25$, and block attackers once they cross this threshold. The accuracy and transferability (of adversarial examples crafted using the test set from $\tilde g$ on to $g$) for each dataset is tabulated in Tables~\ref{tab:accuracy} and \ref{tab:transferability} for attackers with a budget of 100K queries. We make extended results for other budgets and attacks available in Appendix~\ref{apd:first}. The deployment of VarDetect reduces both metrics of the extracted model, demonstrating that the attacks are foiled.

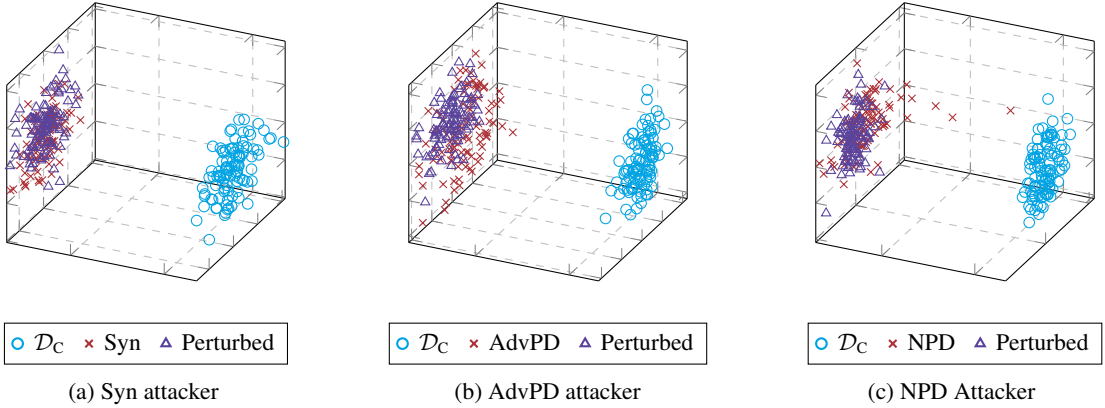
\begin{figure*}[t]
	\centering
	\begin{subfigure}{0.32\linewidth}
		\centering
		\small
		\begin{tikzpicture}
			\begin{axis}[grid=major,grid style={dashed, gray!50},xticklabels={},width=\textwidth,height=\textwidth,yticklabels={},zticklabels={},legend image post style={scale=1.0,thick},
				legend style={/tikz/every even column/.append style={column sep=0.2cm},at={(0.5,-0.15)},
					anchor=north,legend columns=-1}]
				\addplot3+[only marks,mark=o,color=Cerulean] table {latex_data/adaptive_attack/0.001/pos_uniform_uniform-m-pos.txt};
				\addlegendentry{\phantom{.}$\mathcal{D}_\text{C}$}
				\addplot3+[only marks,mark=x,color=Maroon] table {latex_data/adaptive_attack/0.001/pos_uniform_uniform-m-uniform.txt};
				\addlegendentry{\phantom{.}Syn}
				\addplot3+[only marks,mark=triangle,color=Violet] table {latex_data/adaptive_attack/0.001/pos_uniform_uniform-m-uniform-m.txt};
				\addlegendentry{\phantom{.}Perturbed}
			\end{axis}
		\end{tikzpicture}
		\caption{Syn attacker}
	\end{subfigure}
	\begin{subfigure}{0.32\linewidth}
		\centering
		\small
		\begin{tikzpicture}
			\begin{axis}[grid=major,grid style={dashed, gray!50},xticklabels={},width=\textwidth,height=\textwidth,yticklabels={},zticklabels={},legend image post style={scale=1.0,thick},
				legend style={/tikz/every even column/.append style={column sep=0.2cm},at={(0.5,-0.15)},
					anchor=north,legend columns=-1}]
				\addplot3+[only marks,mark=o,color=Cerulean] table {latex_data/adaptive_attack/0.001/pos_jbda-pos.txt};
				\addlegendentry{\phantom{.}$\mathcal{D}_\text{C}$}
				\addplot3+[only marks,mark=x,color=Maroon] table {latex_data/adaptive_attack/0.001/pos_jbda-aug.txt};
				\addlegendentry{\phantom{.}AdvPD}
				\addplot3+[only marks,mark=triangle,color=Violet] table {latex_data/adaptive_attack/0.001/pos_jbda-aug-m.txt};
				\addlegendentry{\phantom{.}Perturbed}
			\end{axis}
		\end{tikzpicture}
		\caption{AdvPD attacker}
	\end{subfigure}
	\begin{subfigure}{0.32\linewidth}
		\centering
		\small
		\begin{tikzpicture}
			\begin{axis}[grid=major,grid style={dashed, gray!50},xticklabels={},width=\textwidth,height=\textwidth,yticklabels={},zticklabels={},legend image post style={scale=1.0,thick},
				legend style={/tikz/every even column/.append style={column sep=0.2cm},at={(0.5,-0.15)},
					anchor=north,legend columns=-1}]
				\addplot3+[only marks,mark=o,color=Cerulean] table {latex_data/adaptive_attack/0.001/pos_imagenet_imagenet-m-pos.txt};
				\addlegendentry{\phantom{.}$\mathcal{D}_\text{C}$}
				\addplot3+[only marks,mark=x,color=Maroon] table {latex_data/adaptive_attack/0.001/pos_imagenet_imagenet-m-imagenet.txt};
				\addlegendentry{\phantom{.}NPD}
				\addplot3+[only marks,mark=triangle,color=Violet] table {latex_data/adaptive_attack/0.001/pos_imagenet_imagenet-m-imagenet-m.txt};
				\addlegendentry{\phantom{.}Perturbed}
			\end{axis}
		\end{tikzpicture}
		\caption{NPD Attacker}
	\end{subfigure}

	\caption{Projection of the latent vectors $\vect{z}$ into 3D (via a PCA projection) for samples $\vect{x}$ drawn from various attacker datasets and those of their corresponding perturbed counterparts (generated by white-box gradient adaptive attackers), obtained after 500 iterations of FGSM with a step size of $\epsilon = 0.001$, compared to latent vectors of confidential dataset samples. Plots are generated for the MNIST dataset.}
	\label{fig:adaptivez}
\end{figure*}

\subsection{Adaptive Attackers}
\label{sec:adaptive}
\cite{adaptive} demonstrate that defenses which are evaluated only against known attacks could fail with simple adaptations that exploit knowledge about the defense. For a comprehensive evaluation of VarDetect in situations where insider knowledge has been compromised, we consider two adaptive attackers:

\begin{figure}[t]
	\centering
	\hfill\begin{subfigure}{0.45\linewidth}
		\begin{tikzpicture}
			\begin{axis}[
				legend style={at={(0.5,-0.55)},
					anchor=north,legend columns=-1,font=\tiny},
				axis x line=left,
				axis y line=left,
				bar width=0.2cm,
				yticklabels={},
				enlarge y limits=0.10,
				extra y ticks=0.25,
				extra y tick labels={0.25},
				extra y tick style={
					grid style={
						black,
						dashed,
						/pgfplots/on layer=axis foreground,
					},
				},
				width=\linewidth,
				height=0.55\linewidth
				]
				\addplot[very thick,color=Maroon,dotted] table[x=x,y=y,col sep=space] {latex_data/vae/mmd/fashion-img-0.05.txt};
				\addplot[very thick,color=Purple,loosely dotted] table[x=x,y=y,col sep=space] {latex_data/vae/mmd/fashion-unif-0.05.txt};
				\addplot[thick,color=orange] table[x=x,y=y,col sep=space] {latex_data/vae/mmd/fashion-test-0.05.txt};
				\addplot[thick,color=Cerulean,dashed] table[x=x,y=y,col sep=space] {latex_data/vae/mmd/fashion-pap-0.05.txt};
			\end{axis}
		\end{tikzpicture}
		\caption{Spaced-out attacker: 5\%}
	\end{subfigure}\hfill
	\begin{subfigure}{0.45\linewidth}
		\begin{tikzpicture}
			\begin{axis}[
				legend style={at={(0.5,-0.55)},
					anchor=north,legend columns=-1,font=\tiny},
				axis x line=left,
				axis y line=left,
				bar width=0.2cm,
				enlarge y limits=0.10,
				yticklabels={},
				extra y ticks=0.25,
				extra y tick labels={0.25},
				extra y tick style={
					grid style={
						black,
						dashed,
						/pgfplots/on layer=axis foreground,
					},
				},
				width=\linewidth,
				height=0.55\linewidth
				]
				\addplot[very thick,color=Maroon,dotted] table[x=x,y=y,col sep=space] {latex_data/vae/mmd/fashion-img-0.15.txt};
				\addplot[very thick,color=Purple,loosely dotted] table[x=x,y=y,col sep=space] {latex_data/vae/mmd/fashion-unif-0.15.txt};
				\addplot[thick,color=orange] table[x=x,y=y,col sep=space] {latex_data/vae/mmd/fashion-test-0.15.txt};
				\addplot[thick,color=Cerulean,dashed] table[x=x,y=y,col sep=space] {latex_data/vae/mmd/fashion-pap-0.15.txt};
			\end{axis}
		\end{tikzpicture}
		\caption{Spaced-out attacker: 15\%}
	\end{subfigure}\hfill\\
	
	\caption{MMD computed by VarDetect over time (F-MNIST dataset) for spaced-out attackers at different dilutions. (PD \ref{benign_mmd}, Syn \ref{unif_mmd}, AdvPD \ref{pap_mmd}, NPD \ref{img_mmd})}
	\label{fig:mmdovertime2}
\end{figure}
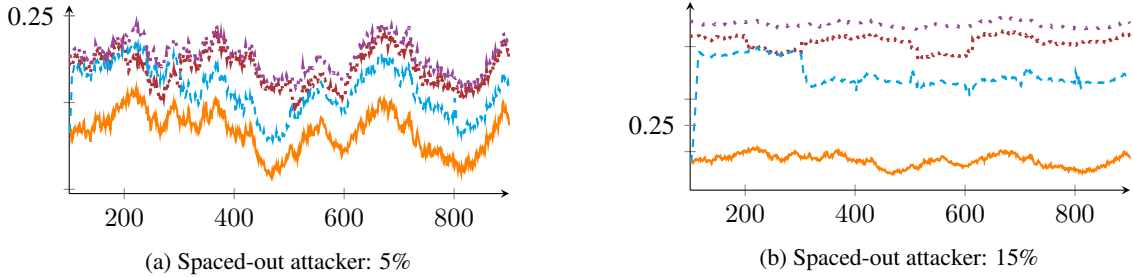

\subsubsection{Spaced-out Adaptive Attackers}
\label{sec:spacedout}
We first consider an attacker that is aware of the stateful nature of VarDetect, and its query buffering strategy. Such an attacker can intelligently space out its $\mathcal{D}_\text{A}$ queries over time, and, in the interim, fire innocuous queries from a benign distribution, e.g., from a limited dataset of PD samples. Let the \textit{dilution factor} of an attacker be the rate at which it fires malicious queries, e.g., an attacker with dilution factor $10\%$ fires 10 attacker queries, followed by 90 benign queries. Note that our spaced-out attacker also addresses the scenario in which one or more individuals who share a single MLaaS user account are malicious (but not all).

We consider two sets of such attackers extracting F-MNIST at dilution factors of 5\% and 15\% respectively, and plot the corresponding MMD over time in Figure~\ref{fig:mmdovertime2}. The attackers with a dilution factor of $15\%$ are detected using a threshold of $\delta = 0.25$, while the attackers using a dilution factor of $5\%$ are not. By titrating the threshold value $\delta$, the security team can make VarDetect sensitive to various values of MMD over time. As the threshold is lowered, the attacker is forced to decrease their dilution factor, at the cost of increased query complexity.

We present extended results for the spaced-out attacker in Appendix~\ref{apd:first}, for dilution factors of 5\%, 15\% and 25\% across all three confidential datasets. We summarize our findings as follows: using a threshold of $\delta = 0.25$ allows us to detect all attackers at a dilution factor of 15\%, except for the NPD attacker on the GTSR dataset. This, too, is detected at a dilution factor of 25\%. Consequently, the overhead of having to fire innocuous queries is increased by $4\times$-$6.67\times$.

\subsubsection{White-Box Gradients Attackers}
\label{sec:evasion}

Next, we consider an adaptive attacker that is motivated to perturb attacker samples $\vect{x} \in \mathcal{D}_\text{A}$ to form modified $\vect{\tilde x} = \vect{x} + \Delta$ such that they are more likely to pass for benign samples, i.e., the encoding of $\vect{\tilde x}$ is closer to the encodings of $\mathcal{D}_\text{C}$ samples than it is to the encodings of $\mathcal{D}_\text{O}$ samples. We assume that this adaptive attacker has white-box access to the VAE encoder $f_\mu$, as well as knowledge of the values of $\vect{\vect{\mu}_\text{O}}$, $\vect{\vect{\mu}_\text{C}}$, $\vect{\sigma_\text{O}}$ and $\vect{\sigma_\text{C}}$ chosen by the MLaaS service provider.

\paragraph{Attack Method}
Starting with a sample $\vect{x} \in \mathcal{D}_\text{A}$, our attacker uses iterative FGSM to nudge its latent encoding in the direction $\vect{\mu}_\text{C}$, with the intent of avoiding detection:
\begin{equation*}
	\text{\small Repeatedly: } \;\;\; \vect{x} \gets \vect{x} + \epsilon \sgn(\nabla_\vect{x} \lVert f_\mu(\vect{x}) - \vect{\vect{\mu}_\text{C}} \rVert)
\end{equation*}

We perform this experiment for all three sets of attackers against a classification model trained on the MNIST dataset of \cite{mnist}, and plot 3D projections of the initial and final latent space encodings after the attack in Figure~\ref{fig:adaptivez} (after $500$ iterations using $\epsilon = 0.001$). As is evident from these figures, our adaptive attacker fails to use iterative FGSM method to avoid detection, and the perturbed samples are detected at a threshold of $\delta = 0.25$.

We perform a grid search by varying {$\epsilon$ in the range $\epsilon \in \{1, 10^{-1}, 10^{-2}, 10^{-3}, 10^{-4}\}$}, and run each instance up to $5000$ iterations of FGSM. Note that $\epsilon$ may be viewed either as the step size (holding a constant learning rate of $1$), or as the learning rate (holding step size constant as $1$). In no case does the attack succeed.

\section{Conclusion}

In this work, we design VarDetect: a framework to detect model extraction attacks targeted at MLaaS providers, by continuously monitoring the queries made by each user to it. VarDetect requires no access to attacker data, and it is the first detection mechanism that raises an alarm for all three types of model extraction attacks in the literature. We demonstrate that with VarDetect deployed as an automated defense mechanism, the task accuracy of extracted substitute models is reduced. Finally, VarDetect is demonstrated to hold up against two types of adaptive attackers: either halting extraction altogether, or increasing the overhead of innocuous queries.

\vfill
\pagebreak

\bibliographystyle{unsrtnat}
\bibliography{references}

\vfill
\pagebreak

\appendix

\section{Additional Model Extraction Attacks}\label{apd:first}

In this section, we demonstrate the efficacy of VarDetect against a larger suite of attacks. We consider the following classes of attacks:

\subsection{Synthetic Attacks (Syn)}
In addition to the Uniform Retraining attack of \citet{tramer}, we additionally evaluate their Line Search Retraining attack, using the publicly available implementation\footnote{https://github.com/ftramer/Steal-ML/}. As no implementation of Adaptive Retraining is available for their experiments on neural networks, we omit this attack.

\subsection{Adversarial Problem Domain Attacks (AdvPD)}
In addition to the JSMA attack of \citet{papernot}, we also evaluate the following additional attacks introduced by \citet{prada}:
\begin{itemize}
	\item \textbf{Non-targeted FGSM (N FGSM)}, which uses the FGSM method of adversarial example generation proposed by \cite{fgsm}, to perturb $x$ away from its original label $y$ ($\mathcal{L}$ being the classifier loss function): $$\vect{\tilde{x}} \gets \vect{x} + \epsilon \cdot \sgn(\nabla_\vect{x} \mathcal{L}(y, S(\vect{x})))$$
	\item \textbf{Non-targeted iterative FGSM (N I-FGSM)}, where iterative FGSM of \cite{ifgsm} is used in lieu of FGSM (the objective remaining unchanged from the NF attack)
	\item \textbf{Targeted FGSM (T-RND FGSM)}, which uses FGSM to perturb $x$ towards a target class $y_r$ chosen uniformly at random from among all possible classes: $$\vect{\tilde{x}} \gets \vect{x} - \epsilon \cdot \sgn(\nabla_\vect{x} \mathcal{L}(y_r, S(\vect{x})))$$
	\item \textbf{Targeted iterative FGSM (T-RND I-FGSM)}, where iterative FGSM is used in lieu of FGSM (the objective remaining the same as that of the TF attack)
\end{itemize}
We implement these attacks, using these update rules, as described in the original paper by \cite{papernot}.

\subsection{Non-Problem Domain Attacks (NPD)}
In addition to the Adversarial + K-Center strategy outlined in the paper, we consider with the following additional strategies proposed by \cite{activethief}:
\begin{itemize}
	\item \textbf{Random}, where the samples $\vect{x}$ to be queried are chosen uniformly at random
	\item \textbf{Uncertainty}, where samples $\vect{x}$ with the highest entropy of the predicted probability vector $S(\vect{x})$ are chosen, as in \cite{uncertainty}, where entropy is calculated as:
	\begin{equation*}
		\mathcal{H}\big(S(\vect{x})\big) = \textstyle\sum\limits_j \log S_j (\vect{x})
	\end{equation*}
	where $S_j(\vect{x})$ is the $j$\textsuperscript{th} component of the vector $S_(\vect{x})$.
	\item \textbf{DeepFool-based Active Learning (DFAL)}, where samples which lie close to the decision boundary are chosen. The DeepFool technique of \cite{deepfool} is used on samples $\vect{x}$ to obtain adversarial $\vect{\tilde x}$. $\alpha = \lVert \vect{x} - \vect{\tilde x} \rVert$ is computed, and samples with the lowest $\alpha$ values are chosen, following the method outlined by \cite{dfal}.
	\item \textbf{k-Center}, where diverse samples are selected by choosing samples that lie farthest apart in an Euclidean distance sense, as in \cite{coreset}.
\end{itemize}
Our implementations of these attacks are based on their public implementations \footnote{https://bitbucket.org/iiscseal/activethief/}.

\subsection{Model Accuracy and Transferability}

We calculate the test accuracy (\%) and transferability success rate (\%) of adversarial examples when using a threshold of $\delta = 0.25$ to automatically block attackers, and present the results in Tables~\ref{tab:accuracy2} and ~\ref{tab:transferability2} respectively.
As before, the substitute model task accuracy for models extracted from a defended model is lower than those extracted from undefended models and the transferability success rate is almost always lowered when crafting adversarial examples using the extracted model. Thus, our observations from Section~\ref{sec:acctrans} are consistent with a far broader range of attacks and query budgets.

\begin{table*}[t]
	\centering
	\caption{Test set accuracy (\%) of substitute models obtained by model extraction attackers, when the MLaaS model is defended using VarDetect (blocking attacks that cross a MMD threshold of $\delta = 0.5$), compared to when it is not. The results clearly indicate that VarDetect reduces the accuracy of the extracted model, as desired.}
	\label{tab:accuracy2}
	
	\resizebox{\linewidth}{!}{
		\begin{tabular}{lrrrrrrr}
			\toprule
			& Defended & \multicolumn{6}{c}{Undefended} \\ \cmidrule(lr){2-2} \cmidrule(lr){3-8}
			\multicolumn{1}{l}{Attacker query budget $\rightarrow$} & \multicolumn{1}{c}{\phantom{xxx}$\infty$} & \multicolumn{1}{c}{10K} & \multicolumn{1}{c}{15K} & \multicolumn{1}{c}{20K} & \multicolumn{1}{c}{25K} & \multicolumn{1}{c}{30K} & \multicolumn{1}{c}{100K}\\
			\midrule
			\multicolumn{8}{c}{\textsc{Fashion-MNIST} of \cite{fmnist}}\\\midrule
			Syn-Uniform Retraining & \bf 7.56 & 15.64 & 13.97 & 21.34 & 14.06 & 15.85 & 25.33\\
			Syn-Line Search Retraining & \bf 6.78 & 11.91 & 14.24 & 14.31 & 12.94 & 15.28 & 20.47\\
			AdvPD JSMA & \bf 75.50 & 80.86 & 79.65 & 82.61 & 81.53 & 83.92 & 84.66\\
			AdvPD N FGSM & \bf 74.37 & 79.27 & 79.11 & 78.57 & 81.15 & 82.92 & 83.63\\
			AdvPD N I-FGSM & \bf 76.61 & 80.93 & 81.72 & 82.90 & 84.07 & 84.35 & 86.19\\
			AdvPD T-RND FGSM & \bf 76.68 & 79.74 & 79.08 & 80.69 & 78.15 & 78.72 & 82.87\\
			AdvPD T-RND I-FGSM & \bf 74.78 & 82.61 & 82.80 & 84.55 & 85.20 & 84.63 & 87.34\\
			NPD-ActiveThief (Random) & \bf 9.96 & 65.06 & 72.89 & 70.92 & 72.07 & 69.67 & 79.64\\
			NPD-ActiveThief (Uncertainty) & \bf 9.99 & 70.33 & 77.94 & 73.99 & 76.36 & 77.76 & 83.15\\
			NPD-ActiveThief (DFAL) & \bf 10.02 & 60.04 & 76.39 & 75.87 & 78.97 & 78.03 & 79.30\\
			NPD-ActiveThief (k-Center) & \bf 10.00 & 77.71 & 74.89 & 79.96 & 78.80 & 81.99 & 82.25\\
			NPD-ActiveThief (DFAL + k-Center) & \bf 12.24 & 71.81 & 75.81 & 81.39 & 79.92 & 80.40 & 80.77\\
			\midrule
			\multicolumn{8}{c}{\textsc{German Traffic Sign Recognition} of \cite{gtsr}}\\\midrule
			Syn-Uniform Retraining & \bf 5.19 & 17.58 & 29.89 & 33.50 & 33.59 & 44.21 & 69.10\\
			Syn-Line Search Retraining & \bf 3.28 & 10.15 & 12.60 & 18.61 & 29.88 & 33.78 & 72.67\\
			AdvPD JSMA & \bf 54.07 & 73.99 & 79.80 & 82.32 & 81.00 & 85.83 & 90.44\\
			AdvPD N FGSM & \bf 58.02 & 71.86 & 74.27 & 75.87 & 77.17 & 77.23 & 83.60\\
			AdvPD N I-FGSM & \bf 60.16 & 65.86 & 67.93 & 72.71 & 68.27 & 64.75 & 80.72\\
			AdvPD T-RND FGSM & \bf 57.43 & 69.02 & 76.57 & 75.91 & 80.45 & 79.93 & 85.07\\
			AdvPD T-RND I-FGSM & \bf 55.75 & 67.34 & 72.59 & 71.48 & 75.90 & 84.80 & 91.11\\
			NPD-ActiveThief (Random) & \bf 7.40 & 53.02 & 63.48 & 64.92 & 68.44 & 68.08 & 85.69\\
			NPD-ActiveThief (Uncertainty) & \bf 6.02 & 54.45 & 61.33 & 69.02 & 74.03 & 74.98 & 83.93\\
			NPD-ActiveThief (DFAL) & \bf 7.59 & 59.39 & 65.58 & 68.95 & 71.59 & 70.44 & 85.42\\
			NPD-ActiveThief (k-Center) & \bf 5.76 & 53.94 & 62.31 & 66.04 & 69.75 & 68.02 & 86.60\\
			NPD-ActiveThief (DFAL + k-Center) & \bf 6.74 & 56.20 & 64.39 & 68.83 & 70.10 & 71.76 & 84.45\\
			\midrule
			\multicolumn{8}{c}{\textsc{StreetView House Numbers} of \cite{svhn}}\\\midrule
			Syn-Uniform Retraining & \bf 14.57 & 16.43 & 37.91 & 38.56 & 39.45 & 45.22 & 50.71\\
			Syn-Line Search Retraining & \bf 11.05 & 15.15 & 24.17 & 27.47 & 40.64 & 45.02 & 54.38\\
			AdvPD JSMA & \bf 41.38 & 57.17 & 59.84 & 59.82 & 65.23 & 65.03 & 71.12\\
			AdvPD N FGSM & \bf 42.77 & 60.14 & 63.84 & 66.97 & 65.62 & 66.86 & 67.83\\
			AdvPD N I-FGSM & \bf 36.71 & 43.62 & 46.72 & 42.02 & 48.32 & 48.08 & 53.15\\
			AdvPD T-RND FGSM & \bf 36.24 & 63.17 & 65.21 & 65.80 & 65.91 & 68.40 & 68.82\\
			AdvPD T-RND I-FGSM & \bf 40.19 & 48.01 & 48.81 & 46.53 & 54.45 & 53.90 & 59.75\\
			NPD-ActiveThief (Random) & \bf 15.42 & 65.30 & 74.34 & 71.78 & 74.32 & 74.18 & 81.93\\
			NPD-ActiveThief (Uncertainty) & \bf 12.50 & 64.75 & 66.79 & 69.59 & 69.31 & 73.71 & 82.43\\
			NPD-ActiveThief (DFAL) & \bf 11.90 & 68.55 & 68.73 & 74.70 & 71.42 & 77.25 & 79.07\\
			NPD-ActiveThief (k-Center) & \bf 15.33 & 67.54 & 67.22 & 73.82 & 76.79 & 77.47 & 82.76\\
			NPD-ActiveThief (DFAL + k-Center) & \bf 15.67 & 67.93 & 70.09 & 72.55 & 74.84 & 77.22 & 82.49\\
			\bottomrule
	\end{tabular}}
\end{table*}

\begin{table*}[t]
	\centering
	\caption{The transferability success rate (\%) of substitute models obtained by model extraction attackers, when the MLaaS model is defended using VarDetect (blocking attacks that cross a MMD threshold of $\delta = 0.5$), compared to when it is not. The results clearly indicate that in most cases, VarDetect reduces the transferability success rate of the extracted model, as desired.}
	\label{tab:transferability2}
	
	\resizebox{\linewidth}{!}{
		\begin{tabular}{lrrrrrrr}
			\toprule
			& Defended & \multicolumn{6}{c}{Undefended} \\ \cmidrule(lr){2-2} \cmidrule(lr){3-8}
			\multicolumn{1}{l}{Attacker query budget $\rightarrow$} & \multicolumn{1}{c}{\phantom{xxx}$\infty$} & \multicolumn{1}{c}{10K} & \multicolumn{1}{c}{15K} & \multicolumn{1}{c}{20K} & \multicolumn{1}{c}{25K} & \multicolumn{1}{c}{30K} & \multicolumn{1}{c}{100K}\\
			\midrule
			\multicolumn{8}{c}{\textsc{Fashion-MNIST} of \cite{fmnist}}\\\midrule
			Syn-Uniform Retraining & \bf 42.35 & 59.56 & 59.14 & 59.31 & 56.67 & 58.17 & 60.74\\
			Syn-Line Search Retraining & \bf 46.12 & 61.48 & 60.30 & 61.92 & 63.35 & 64.49 & 58.96\\
			AdvPD JSMA & 63.13 & \bf 62.22 & 63.38 & 63.41 & 65.74 & 62.22 & 68.73\\
			AdvPD N FGSM & \bf 58.92 & 62.02 & 68.91 & 67.20 & 69.61 & 68.66 & 72.05\\
			AdvPD N I-FGSM & 62.36 & \bf 57.00 & 58.72 & 60.47 & 59.46 & 61.36 & 64.19\\
			AdvPD T-RND FGSM & 64.53 & 72.14 & \bf 64.37 & 66.72 & 65.66 & 66.30 & 71.88\\
			AdvPD T-RND I-FGSM & 64.86 & 62.28 & \bf 61.02 & 61.79 & 61.12 & 60.59 & 65.67\\
			NPD-ActiveThief (Random) & \bf 56.73 & 72.29 & 79.94 & 71.34 & 68.38 & 73.60 & 78.74\\
			NPD-ActiveThief (Uncertainty) & \bf 54.20 & 79.97 & 78.58 & 80.81 & 81.63 & 79.20 & 80.74\\
			NPD-ActiveThief (DFAL) & \bf 54.21 & 75.93 & 72.69 & 78.42 & 82.10 & 77.94 & 83.11\\
			NPD-ActiveThief (k-Center) & \bf 56.83 & 75.50 & 79.17 & 82.05 & 83.71 & 84.47 & 80.35\\
			NPD-ActiveThief (DFAL + k-Center) & \bf 55.79 & 79.63 & 77.91 & 74.93 & 76.99 & 71.16 & 81.10\\
			\midrule
			\multicolumn{8}{c}{\textsc{German Traffic Sign Recognition} of \cite{gtsr}}\\\midrule
			Syn-Uniform Retraining & \bf 13.61 & 64.10 & 68.79 & 72.32 & 73.14 & 75.25 & 84.51\\
			Syn-Line Search Retraining & \bf 33.16  & 54.24 & 59.41 & 66.85 & 73.20 & 74.69 & 88.08\\
			AdvPD JSMA & \bf 68.53 & 72.43 & 74.28 & 73.45 & 76.94 & 78.24 & 80.81\\
			AdvPD N FGSM & \bf 67.79 & 77.64 & 79.28 & 78.27 & 81.20 & 80.10 & 85.57\\
			AdvPD N I-FGSM & 56.96 & 49.98 & 57.14 & 43.79 & \bf 41.99 & 59.78 & 66.00\\
			AdvPD T-RND FGSM & \bf 62.26 & 80.16 & 79.26 & 80.10 & 80.11 & 80.38 & 84.62\\
			AdvPD T-RND I-FGSM & 62.83 & \bf 62.33 & 68.31 & 62.14 & 69.86 & 73.64 & 77.90\\
			NPD-ActiveThief (Random) & \bf 31.16 & 83.50 & 85.76 & 88.67 & 87.88 & 89.61 & 94.56\\
			NPD-ActiveThief (Uncertainty) & \bf 29.28 & 82.43 & 86.67 & 87.58 & 88.32 & 89.93 & 94.28\\
			NPD-ActiveThief (DFAL) & \bf 35.95 & 82.34 & 86.48 & 86.19 & 89.68 & 89.41 & 94.42\\
			NPD-ActiveThief (k-Center) & \bf 24.74 & 85.23 & 87.80 & 84.38 & 89.83 & 90.68 & 92.97\\
			NPD-ActiveThief (DFAL + k-Center) & \bf 37.94 & 84.90 & 80.51 & 88.89 & 90.72 & 90.40 & 93.96\\
			\midrule
			\multicolumn{8}{c}{\textsc{StreetView House Numbers} of \cite{svhn}}\\\midrule
			Syn-Uniform Retraining & \bf 31.54 & 33.55 & 74.85 & 76.64 & 76.02 & 77.02 & 82.51\\
			Syn-Line Search Retraining & \bf 28.12 & 31.71 & 42.29 & 51.03 & 71.08 & 79.35 & 85.70\\
			AdvPD JSMA & \bf 53.52 & 67.69 & 65.74 & 70.96 & 73.90 & 75.09 & 82.30\\
			AdvPD N FGSM & \bf 57.92 & 77.21 & 79.73 & 78.57 & 78.87 & 82.20 & 83.22\\
			AdvPD N I-FGSM & 50.02 & 28.68 & 39.56 & \bf 22.55 & 30.56 & 44.28 & 39.75\\
			AdvPD T-RND FGSM & \bf 52.31 & 77.44 & 80.06 & 80.54 & 80.29 & 82.07 & 82.58\\
			AdvPD T-RND I-FGSM & 54.61 & 46.44 & \bf 42.23 & 39.34 & 53.49 & 55.56 & 57.27\\
			NPD-ActiveThief (Random) & \bf 24.22 & 81.54 & 87.35 & 88.28 & 88.24 & 89.95 & 91.30\\
			NPD-ActiveThief (Uncertainty) & \bf 24.82 & 84.30 & 84.64 & 86.44 & 87.24 & 88.51 & 92.52\\
			NPD-ActiveThief (DFAL) & \bf 25.49 & 84.76 & 87.27 & 87.77 & 85.97 & 88.76 & 92.01\\
			NPD-ActiveThief (k-Center) & \bf 41.81 & 83.39 & 86.02 & 87.27 & 88.64 & 87.95 & 92.11\\
			NPD-ActiveThief (DFAL + k-Center) & \bf 26.87 & 84.57 & 85.19 & 84.42 & 88.89 & 86.52 & 92.57\\
			\bottomrule
	\end{tabular}}
\end{table*}

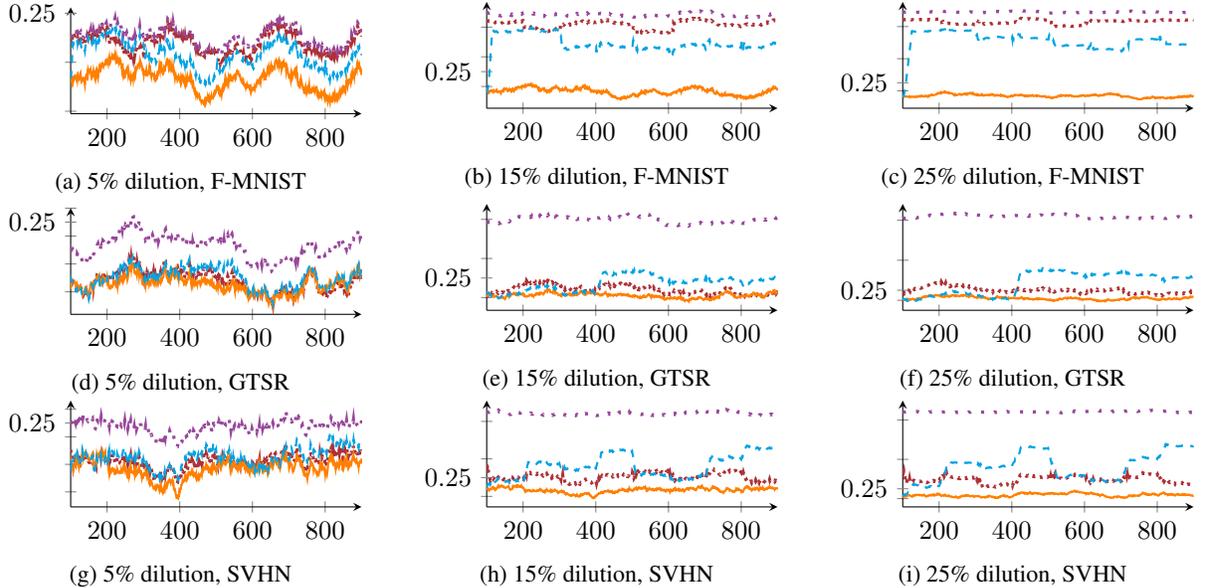
\begin{figure*}[t]
	\centering
	\begin{subfigure}{0.33\linewidth}
		\centering
		\begin{tikzpicture}
			\begin{axis}[
				legend style={at={(0.5,-0.55)},
					anchor=north,legend columns=-1,font=\tiny},
				axis x line=left,
				axis y line=left,
				bar width=0.2cm,
				yticklabels={},
				enlarge y limits=0.10,
				extra y ticks=0.25,
				extra y tick labels={$0.25$},
				extra y tick style={
					grid style={
						black,
						dashed,
						/pgfplots/on layer=axis foreground,
					},
				},
				width=\linewidth,
				height=0.55\linewidth
				]
				\addplot[very thick,color=Maroon,dotted] table[x=x,y=y,col sep=space] {latex_data/vae/mmd/fashion-img-0.05.txt};
				\addplot[very thick,color=Purple,loosely dotted] table[x=x,y=y,col sep=space] {latex_data/vae/mmd/fashion-unif-0.05.txt};
				\addplot[thick,color=orange] table[x=x,y=y,col sep=space] {latex_data/vae/mmd/fashion-test-0.05.txt};
				\addplot[thick,color=Cerulean,dashed] table[x=x,y=y,col sep=space] {latex_data/vae/mmd/fashion-pap-0.05.txt};
			\end{axis}
		\end{tikzpicture}
		\caption{5\% dilution, F-MNIST}
	\end{subfigure}
	\begin{subfigure}{0.33\linewidth}
		\centering
		\begin{tikzpicture}
			\begin{axis}[
				legend style={at={(0.5,-0.55)},
					anchor=north,legend columns=-1,font=\tiny},
				axis x line=left,
				axis y line=left,
				bar width=0.2cm,
				enlarge y limits=0.10,
				yticklabels={},
				extra y ticks=0.25,
				extra y tick labels={$0.25$},
				extra y tick style={
					grid style={
						black,
						dashed,
						/pgfplots/on layer=axis foreground,
					},
				},
				width=\linewidth,
				height=0.55\linewidth
				]
				\addplot[very thick,color=Maroon,dotted] table[x=x,y=y,col sep=space] {latex_data/vae/mmd/fashion-img-0.15.txt};
				\addplot[very thick,color=Purple,loosely dotted] table[x=x,y=y,col sep=space] {latex_data/vae/mmd/fashion-unif-0.15.txt};
				\addplot[thick,color=orange] table[x=x,y=y,col sep=space] {latex_data/vae/mmd/fashion-test-0.15.txt};
				\addplot[thick,color=Cerulean,dashed] table[x=x,y=y,col sep=space] {latex_data/vae/mmd/fashion-pap-0.15.txt};
			\end{axis}
		\end{tikzpicture}
		\caption{15\% dilution, F-MNIST}
	\end{subfigure}
	\begin{subfigure}{0.33\linewidth}
		\centering
		\begin{tikzpicture}
			\begin{axis}[
				legend style={at={(0.5,-0.55)},
					anchor=north,legend columns=-1,font=\tiny},
				axis x line=left,
				axis y line=left,
				bar width=0.2cm,
				enlarge y limits=0.10,
				yticklabels={},
				extra y ticks=0.25,
				extra y tick labels={$0.25$},
				extra y tick style={
					grid style={
						black,
						dashed,
						/pgfplots/on layer=axis foreground,
					},
				},
				width=\linewidth,
				height=0.55\linewidth
				]
				\addplot[very thick,color=Maroon,dotted] table[x=x,y=y,col sep=space] {latex_data/vae/mmd/fashion-img-0.25.txt};
				\addplot[very thick,color=Purple,loosely dotted] table[x=x,y=y,col sep=space] {latex_data/vae/mmd/fashion-unif-0.25.txt};
				\addplot[thick,color=orange] table[x=x,y=y,col sep=space] {latex_data/vae/mmd/fashion-test-0.25.txt};
				\addplot[thick,color=Cerulean,dashed] table[x=x,y=y,col sep=space] {latex_data/vae/mmd/fashion-pap-0.25.txt};
			\end{axis}
		\end{tikzpicture}
		\caption{25\% dilution, F-MNIST}
	\end{subfigure}
	\begin{subfigure}{0.33\linewidth}
		\centering
		\begin{tikzpicture}
			\begin{axis}[
				legend style={at={(0.5,-0.55)},
					anchor=north,legend columns=-1,font=\tiny},
				axis x line=left,
				axis y line=left,
				bar width=0.2cm,
				yticklabels={},
				enlarge y limits=0.10,
				extra y ticks=0.25,
				extra y tick labels={$0.25$},
				extra y tick style={
					grid style={
						black,
						dashed,
						/pgfplots/on layer=axis foreground,
					},
				},
				width=\linewidth,
				height=0.55\linewidth
				]
				\addplot[very thick,color=Maroon,dotted] table[x=x,y=y,col sep=space] {latex_data/vae/mmd/gtsr-img-0.05.txt};
				\addplot[very thick,color=Purple,loosely dotted] table[x=x,y=y,col sep=space] {latex_data/vae/mmd/gtsr-unif-0.05.txt};
				\addplot[thick,color=orange] table[x=x,y=y,col sep=space] {latex_data/vae/mmd/gtsr-test-0.05.txt};
				\addplot[thick,color=Cerulean,dashed] table[x=x,y=y,col sep=space] {latex_data/vae/mmd/gtsr-pap-0.05.txt};
			\end{axis}
		\end{tikzpicture}
		\caption{5\% dilution, GTSR}
	\end{subfigure}
	\begin{subfigure}{0.33\linewidth}
		\centering
		\begin{tikzpicture}
			\begin{axis}[
				legend style={at={(0.5,-0.55)},
					anchor=north,legend columns=-1,font=\tiny},
				axis x line=left,
				axis y line=left,
				bar width=0.2cm,
				enlarge y limits=0.10,
				yticklabels={},
				extra y ticks=0.25,
				extra y tick labels={$0.25$},
				extra y tick style={
					grid style={
						black,
						dashed,
						/pgfplots/on layer=axis foreground,
					},
				},
				width=\linewidth,
				height=0.55\linewidth
				]
				\addplot[very thick,color=Maroon,dotted] table[x=x,y=y,col sep=space] {latex_data/vae/mmd/gtsr-img-0.15.txt};
				\addplot[very thick,color=Purple,loosely dotted] table[x=x,y=y,col sep=space] {latex_data/vae/mmd/gtsr-unif-0.15.txt};
				\addplot[thick,color=orange] table[x=x,y=y,col sep=space] {latex_data/vae/mmd/gtsr-test-0.15.txt};
				\addplot[thick,color=Cerulean,dashed] table[x=x,y=y,col sep=space] {latex_data/vae/mmd/gtsr-pap-0.15.txt};
			\end{axis}
		\end{tikzpicture}
		\caption{15\% dilution, GTSR}
	\end{subfigure}
	\begin{subfigure}{0.33\linewidth}
		\centering
		\begin{tikzpicture}
			\begin{axis}[
				legend style={at={(0.5,-0.55)},
					anchor=north,legend columns=-1,font=\tiny},
				axis x line=left,
				axis y line=left,
				bar width=0.2cm,
				enlarge y limits=0.10,
				yticklabels={},
				extra y ticks=0.25,
				extra y tick labels={$0.25$},
				extra y tick style={
					grid style={
						black,
						dashed,
						/pgfplots/on layer=axis foreground,
					},
				},
				width=\linewidth,
				height=0.55\linewidth
				]
				\addplot[very thick,color=Maroon,dotted] table[x=x,y=y,col sep=space] {latex_data/vae/mmd/gtsr-img-0.25.txt};
				\addplot[very thick,color=Purple,loosely dotted] table[x=x,y=y,col sep=space] {latex_data/vae/mmd/gtsr-unif-0.25.txt};
				\addplot[thick,color=orange] table[x=x,y=y,col sep=space] {latex_data/vae/mmd/gtsr-test-0.25.txt};
				\addplot[thick,color=Cerulean,dashed] table[x=x,y=y,col sep=space] {latex_data/vae/mmd/gtsr-pap-0.25.txt};
			\end{axis}
		\end{tikzpicture}
		\caption{25\% dilution, GTSR}
	\end{subfigure}
	\begin{subfigure}{0.33\linewidth}
		\centering
		\begin{tikzpicture}
			\begin{axis}[
				legend style={at={(0.5,-0.55)},
					anchor=north,legend columns=-1,font=\tiny},
				axis x line=left,
				axis y line=left,
				bar width=0.2cm,
				yticklabels={},
				enlarge y limits=0.10,
				extra y ticks=0.25,
				extra y tick labels={$0.25$},
				extra y tick style={
					grid style={
						black,
						dashed,
						/pgfplots/on layer=axis foreground,
					},
				},
				width=\linewidth,
				height=0.55\linewidth
				]
				\addplot[very thick,color=Maroon,dotted] table[x=x,y=y,col sep=space] {latex_data/vae/mmd/svhn-img-0.05.txt};
				\addplot[very thick,color=Purple,loosely dotted] table[x=x,y=y,col sep=space] {latex_data/vae/mmd/svhn-unif-0.05.txt};
				\addplot[thick,color=orange] table[x=x,y=y,col sep=space] {latex_data/vae/mmd/svhn-test-0.05.txt};
				\addplot[thick,color=Cerulean,dashed] table[x=x,y=y,col sep=space] {latex_data/vae/mmd/svhn-pap-0.05.txt};
			\end{axis}
		\end{tikzpicture}
		\caption{5\% dilution, SVHN}
	\end{subfigure}
	\begin{subfigure}{0.33\linewidth}
		\centering
		\begin{tikzpicture}
			\begin{axis}[
				legend style={at={(0.5,-0.55)},
					anchor=north,legend columns=-1,font=\tiny},
				axis x line=left,
				axis y line=left,
				bar width=0.2cm,
				enlarge y limits=0.10,
				yticklabels={},
				extra y ticks=0.25,
				extra y tick labels={$0.25$},
				extra y tick style={
					grid style={
						black,
						dashed,
						/pgfplots/on layer=axis foreground,
					},
				},
				width=\linewidth,
				height=0.55\linewidth
				]
				\addplot[very thick,color=Maroon,dotted] table[x=x,y=y,col sep=space] {latex_data/vae/mmd/svhn-img-0.15.txt};
				\addplot[very thick,color=Purple,loosely dotted] table[x=x,y=y,col sep=space] {latex_data/vae/mmd/svhn-unif-0.15.txt};
				\addplot[thick,color=orange] table[x=x,y=y,col sep=space] {latex_data/vae/mmd/svhn-test-0.15.txt};
				\addplot[thick,color=Cerulean,dashed] table[x=x,y=y,col sep=space] {latex_data/vae/mmd/svhn-pap-0.15.txt};
			\end{axis}
		\end{tikzpicture}
		\caption{15\% dilution, SVHN}
	\end{subfigure}
	\begin{subfigure}{0.33\linewidth}
		\centering
		\begin{tikzpicture}
			\begin{axis}[
				legend style={at={(0.5,-0.55)},
					anchor=north,legend columns=-1,font=\tiny},
				axis x line=left,
				axis y line=left,
				bar width=0.2cm,
				enlarge y limits=0.10,
				yticklabels={},
				extra y ticks=0.25,
				extra y tick labels={$0.25$},
				extra y tick style={
					grid style={
						black,
						dashed,
						/pgfplots/on layer=axis foreground,
					},
				},
				width=\linewidth,
				height=0.55\linewidth
				]
				\addplot[very thick,color=Maroon,dotted] table[x=x,y=y,col sep=space] {latex_data/vae/mmd/svhn-img-0.25.txt};\label{img_mmd}
				\addplot[very thick,color=Purple,loosely dotted] table[x=x,y=y,col sep=space] {latex_data/vae/mmd/svhn-unif-0.25.txt};\label{unif_mmd}
				\addplot[thick,color=orange] table[x=x,y=y,col sep=space] {latex_data/vae/mmd/svhn-test-0.25.txt};\label{benign_mmd}
				\addplot[thick,color=Cerulean,dashed] table[x=x,y=y,col sep=space] {latex_data/vae/mmd/svhn-pap-0.25.txt};\label{pap_mmd}
			\end{axis}
		\end{tikzpicture}
		\caption{25\% dilution, SVHN}
	\end{subfigure}
	\caption{Change in MMD for different datasets when an adaptive attacker fires queries at different dilution fractions (the plot colors and patterns follow the same conventions as before, i.e., PD \ref{benign_mmd}, Syn \ref{unif_mmd}, AdvPD \ref{pap_mmd} and NPD \ref{img_mmd}).}
	\label{fig:mmdovertime3}
\end{figure*}

\subsection{Spaced-out Adaptive Attacker}

In Figure~\ref{fig:mmdovertime3}, we present extended results for the spaced-out attacker we present in Section~\ref{sec:spacedout}. A threshold of $\delta = 0.5$ is adequate to detect all three types of attacks at dilutions of $15\%$ or above, with the sole exception of the NPD attack on the GTSR dataset. The NPD attacker for the GTSR dataset is detected at a threshold of 25\%, as shown.

\section{Configuration and Reproducibility}\label{apd:second}
Our deep learning models are implemented in Python 2.7.17, using the TensorFlow 1.14 framework of \cite{tensorflow}, and are executed on an NVIDIA GPU using CUDA 10.0 and NVIDIA cuDNN 7.6.4. We additionally use a number of Python packages specified in the \texttt{requirements.txt} file of our public code repository.

We perform our experiments on a server with a 20-core Intel(R) Xeon(R) CPU E5-2630 v4 @ 2.20GHz processor, 32 GB of system memory and equipped with a Titan X (Pascal) GPU accelerator with 12GB of vRAM, running on the Ubuntu 18.04.5 LTS (Bionic Beaver) operating system.

In all of our experiments, we use a program-wide seeds to ensure reproducibility. However, due to the underlying nature of cuDNN-reliant operations such as \texttt{tf.reduce\_sum}, GPU non-determinism may cause the weights of the trained models to change across multiple runs, even on the same system. To this end, we apply the TensorFlow Determinism patch made available by NVIDIA\footnote{https://github.com/NVIDIA/framework-determinism} to reduce GPU non-determinism and aid in reproducibility. We tested our experiments across a number of different servers, and found the results to be consistent within a margin of error. Further information about our seeds is available in the public repository we release as part of this work.

\end{document}